\documentclass[11pt]{article}

\usepackage[preprint]{acl}

\usepackage{times}
\usepackage{latexsym}
\usepackage[T1]{fontenc}
\usepackage[utf8]{inputenc}
\usepackage{microtype}
\usepackage{inconsolata}

\usepackage{graphicx}
\usepackage{booktabs}
\usepackage{amsmath}
\usepackage{amssymb}
\usepackage{amsfonts}
\usepackage{algorithm}
\usepackage{algorithmic}
\usepackage{multirow}
\usepackage[table]{xcolor}
\usepackage{subcaption}
\usepackage{placeins}
\usepackage{enumitem}

\definecolor{Gain}{RGB}{0,128,0}
\definecolor{Drop}{RGB}{200,0,0}

\title{ADHint: Adaptive Hints with Difficulty Priors for Reinforcement Learning}

\author{
	Feng Zhang$^{1,2,*}$ \quad
	Zezhong Tan$^{2,3,*}$ \quad
	Xinhong Ma$^{2}$ \quad
	Ziqiang Dong$^{2,\dagger}$ \\
	\textbf{Xi Leng}$^{2,4}$ \quad
	\textbf{Jianfei Zhao}$^{1,5}$ \quad
	\textbf{Xin Sun}$^{1,\dagger}$ \quad
	\textbf{Yang Yang}$^{2}$ \quad
	\textbf{Guanjun Jiang}$^{2}$ \\
	\vspace{-0.8em}
	\\
	$^{1}$Beijing Institute of Technology \quad
	$^{2}$Qwen Business Unit of Alibaba  \\
	$^{3}$Peking University \quad
	$^{4}$The Chinese University of Hong Kong, Shenzhen \quad
	$^{5}$Zhongguancun Academy \\
	\texttt{\{bit\_zhangfeng, zhqingan, sunxin\}@bit.edu.cn} \\
	\texttt{\{xinhong.mxh, ziqiang.dzq, chris.yang, guanj.jianggj\}@alibaba-inc.com} \\
	\texttt{tzz@stu.pku.edu.cn} \quad
	\texttt{xileng@link.cuhk.edu.cn}
}

\begin{document}
\maketitle

\begingroup
\renewcommand{\thefootnote}{\fnsymbol{footnote}}
\footnotetext[1]{Equal contribution.}
\footnotetext[2]{Corresponding authors.}
\endgroup
\begin{abstract}
To address the limited capability expansion and low sample efficiency of Reinforcement Learning (RL), recent methods have integrated ``hints'' into post-training, which are prefix segments of complete reasoning trajectories, aiming for powerful knowledge expansion and reasoning generalization.
However, existing hint-based RL methods often neglect the role of difficulty in the hint-ratio schedule and relative-advantage estimation, resulting in unstable learning and excessive imitation of off-policy hints.
To address this, we propose ADHint, which explicitly integrates difficulty into both processes to achieve a better trade-off between exploration and imitation.
Specifically, we propose \textbf{Adaptive Hint with Sample Difficulty Prior}, which evaluates the difficulty of each sample under the current policy to schedule an appropriate hint ratio for rollout generation. 
Furthermore, we introduce \textbf{Consistency-based Gradient Modulation} alongside \textbf{Selective Masking for Hint Preservation}, which jointly modulate token-level gradients within hints to prevent biased and destructive updates. 
Additionally, we propose \textbf{Advantage Estimation with Rollout Difficulty Posterior}, which leverages the relative difficulty of rollouts with and without hints to compute their respective advantages, yielding more balanced updates.
Extensive experiments across diverse modalities, scales, model families, and domains show that ADHint achieves superior reasoning capabilities and out-of-distribution generalization. 
Code will be released upon paper acceptance.
\end{abstract}

\section{Introduction}
\label{sec:intro}
\begin{figure}[!t]
    \centering
    \captionsetup{skip=4pt}
    \captionsetup[subfigure]{skip=2pt,justification=centering}
    \begin{subfigure}[t]{0.49\linewidth}
        \centering
        \includegraphics[width=\linewidth]{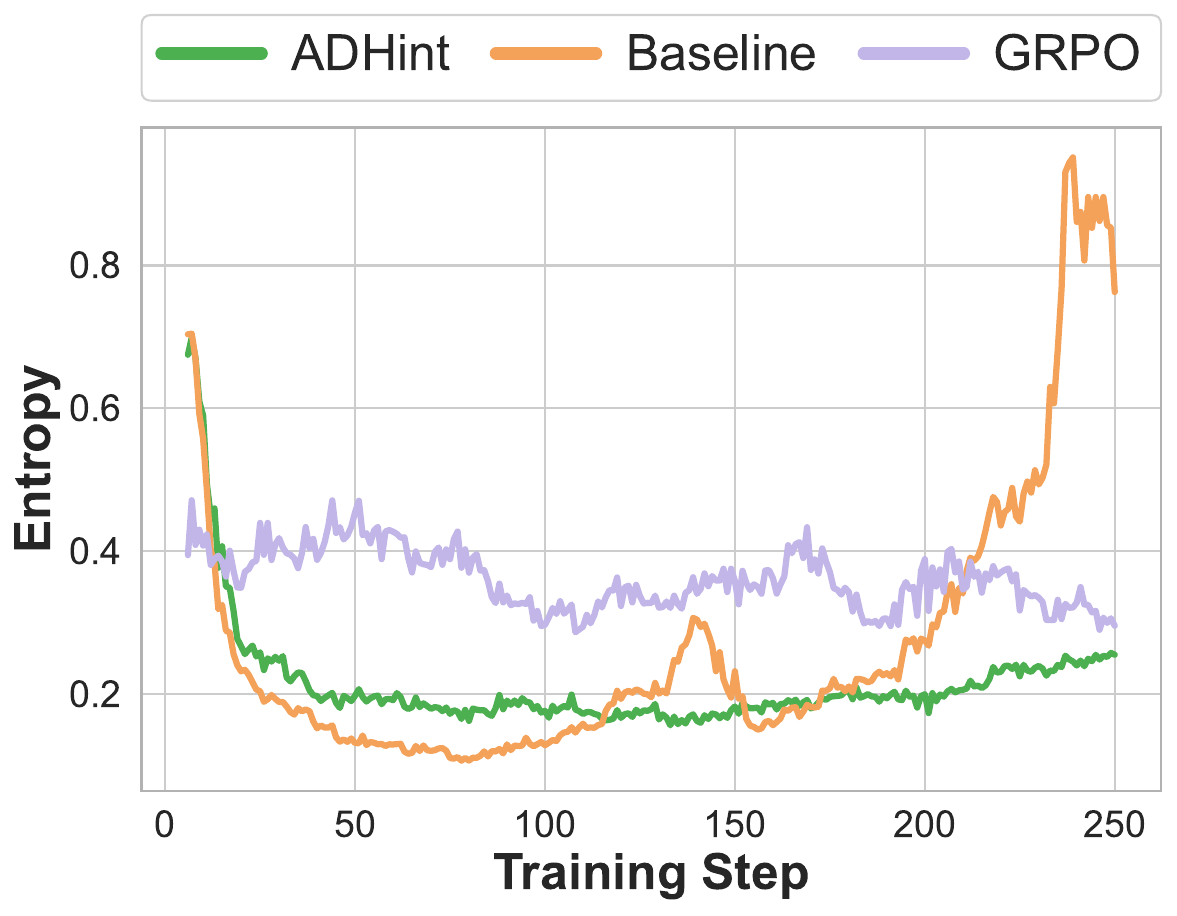}
        \caption{ADHint w/\protect\\annealing hint ratio}
        \label{fig1:sub_a}
    \end{subfigure}
    \hfill
    \begin{subfigure}[t]{0.49\linewidth}
        \centering
        \includegraphics[width=\linewidth]{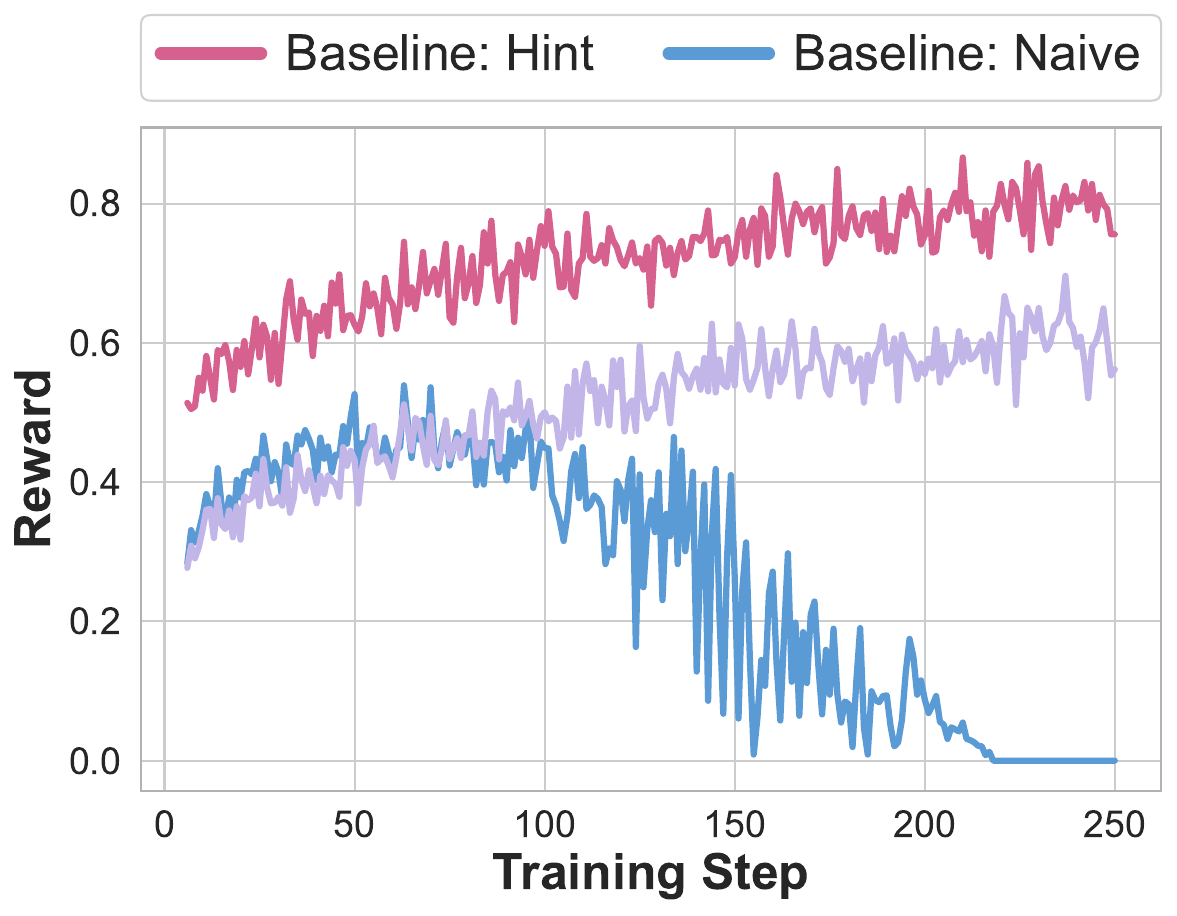}
        \caption{ADHint w/\protect\\standard advantages}
        \label{fig1:sub_b}
    \end{subfigure}

    \caption{
    \textbf{Baseline behavior with existing hint-based designs.}
    ``ADHint w/ annealing hint ratio'' replaces our sample difficulty prior-based hint-ratio schedule (AH-SDP) with a commonly used time-varying schedule.
    ``ADHint w/ standard advantages'' replaces our rollout difficulty posterior-based advantage estimation (AE-RDP) with the standard estimation that pools hint-rollouts and naive-rollouts into a single GRPO group.
    In each subfigure, the \textit{Baseline} curve matches the variant in the title, and ``Baseline: Hint/Naive'' reports statistics from hint- or naive-rollouts only.
    }
    \label{fig:1}
    \vspace{-1.5em}
\end{figure}

\begin{figure*}[!t]
	\centering
	\captionsetup{skip=4pt}
	\captionsetup[subfigure]{skip=2pt}
	
	\begin{subfigure}[b]{0.40\textwidth}
		\centering
		\includegraphics[width=\linewidth]{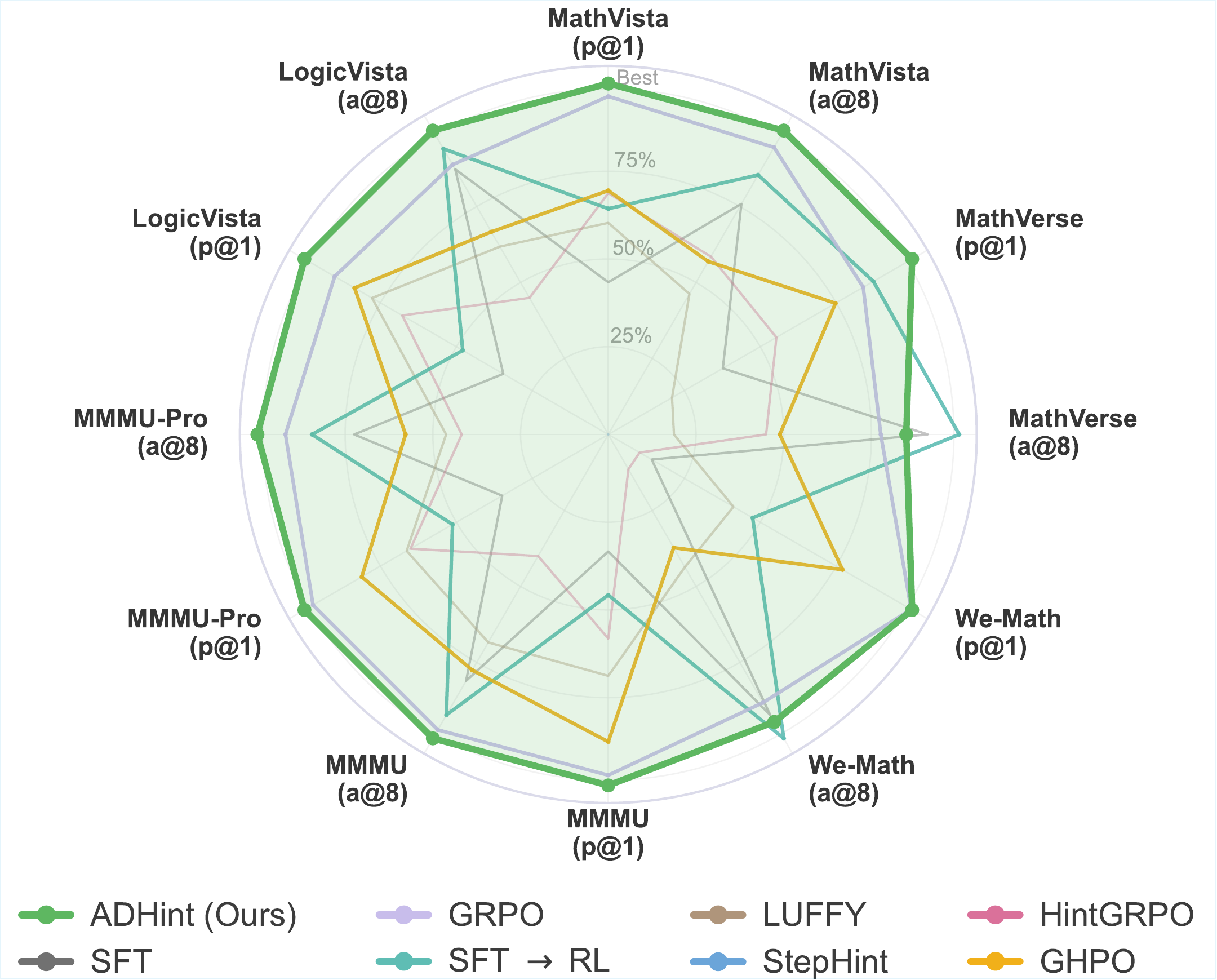}
		\caption{Performance on Qwen2.5-VL-7B.}
	\end{subfigure}%
	\hfill
	\begin{subfigure}[b]{0.58\textwidth}
		\centering
		\includegraphics[width=\linewidth]{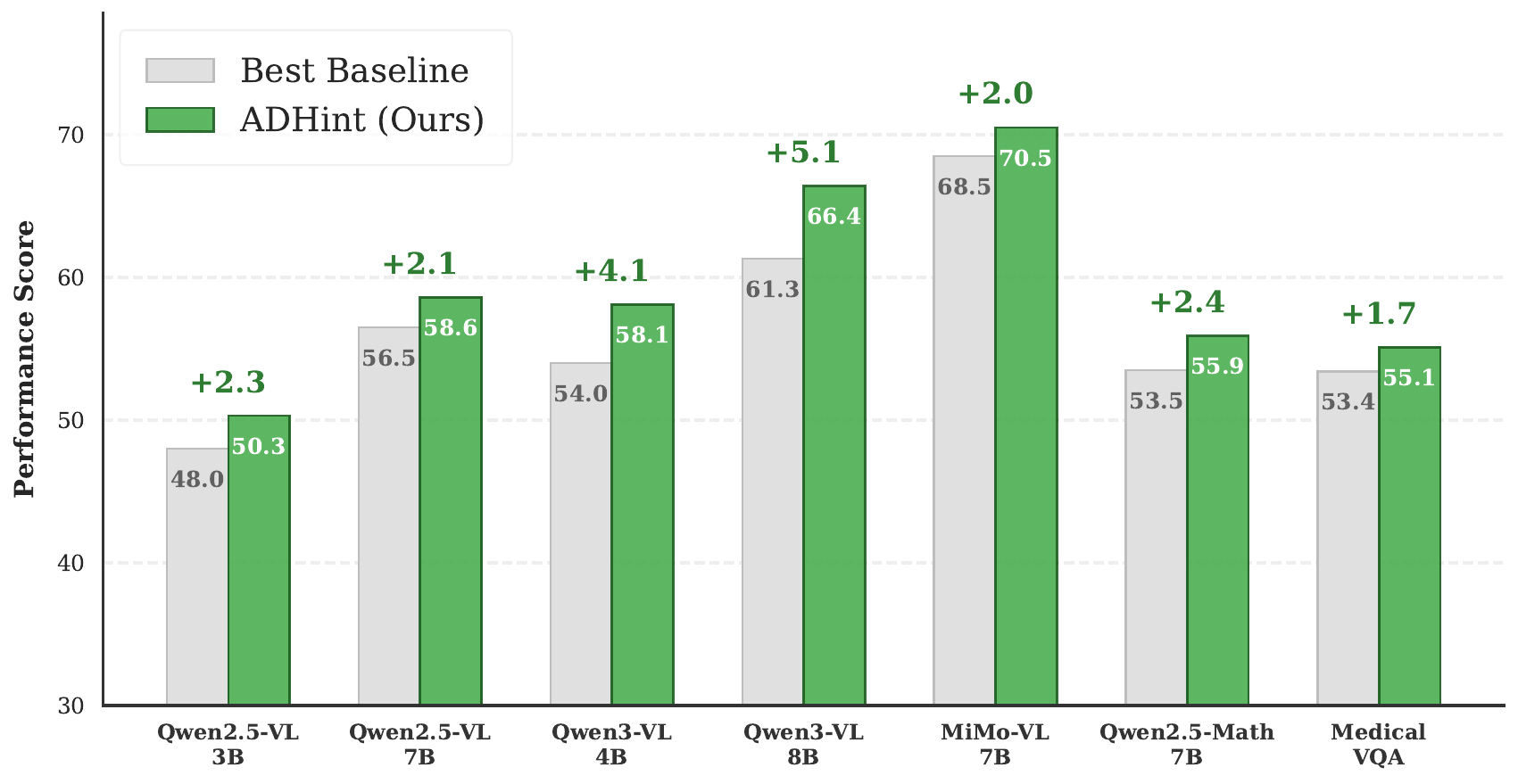}
		\caption{Gains across scenarios.}
	\end{subfigure}
	
	\caption{
		\textbf{Performance Comparison and Generalization.} 
		(a) Radar chart on Qwen2.5-VL-7B-Instruct. Each axis represents a dataset-metric pair (p@1: \textit{pass@1}, a@8: \textit{avg@8}), where ADHint consistently achieves the best or highly competitive performance. 
		(b) Absolute gains over the best baselines. ADHint demonstrates robust improvements across diverse modalities, model scales, model families, and domains.
	}
	\label{fig:2}
	\vspace{-1.2em}
\end{figure*}
Recent advancements in Large Language Models \cite{yang2025qwen3,zeng2025glm} and Multimodal Large Language Models \cite{bai2025qwen3vltechnicalreport,huang2025vision} highlight the potential to enhance reasoning capabilities by large-scale 
Reinforcement Learning with Verifiable Rewards (RLVR) \cite{guo2025deepseek}, 
such as Proximal Policy Optimization (PPO) \cite{schulman2017proximal} and Group Relative Policy Optimization (GRPO) \cite{shao2024deepseekmath}.
However, current on-policy RLVR faces two major challenges: (1) \emph{Limited capability expansion}: RLVR is inherently bounded by the base model. It primarily amplifies existing behaviors and refines known reasoning chains, rather than instilling genuinely novel reasoning abilities beyond its initial capability boundaries~\cite{chusft,li2025system}. 
(2) \emph{Low sample efficiency}: The learning process is bottlenecked by the current policy's performance, yielding critically sparse reward signals that render hard samples difficult to exploit~\cite{wu2025templaterlstructuredtemplateguidedreinforcement,xi2025bapostabilizingoffpolicyreinforcement}.

To mitigate these limitations, recent methods~\cite{Huang_2025_ICCV, liu2025uft} introduce \emph{hints}, defined as prefix segments of complete reasoning trajectories, which guide the policy to explore the continuation, thereby acquiring novel capabilities and solving challenging problems.
Specifically, these methods assign fixed \cite{zhang2025stephint} or time-varying \cite{liu2025uft,huang2025blending} hint ratios to all samples (or only to the hardest samples \cite{Huang_2025_ICCV,liu2025ghpo}),
and then utilize \emph{hint-rollouts} derived from hint-guided reasoning together with \emph{naive-rollouts} generated without hints
for group relative advantage estimation \cite{zhang2025stephint, huang2025blending}.

Despite their empirical successes, existing hint-based RL methods are sensitive to sample-difficulty in the hint-ratio schedule and rollout-difficulty in the relative-advantage estimation.
Specifically, most previous methods apply a uniform hint ratio to both easy and hard samples for model reasoning,
and then generate hint-rollouts with various difficulties. 
Policy learning from such heterogeneous rollouts could induce high-variance model optimization and undermine training stability \cite{bengio2009curriculum,parashar2025curriculum,li2025adacurl}.
As shown in Figure \ref{fig1:sub_a}, the
model with annealing hint ratios (denoted as Baseline) suffers training collapse as its entropy rises sharply late in training, indicating insufficient absorption of external knowledge and a drift in the model's own policy.
In contrast, when incorporating the hint-ratio schedule with sample-difficulty (our method, discussed later), the entropy converges stably, ensuring that the model internalizes robust reasoning abilities rather than superficial patterns within off-policy hints.

As for relative-advantage estimation, most existing approaches pool hint-rollouts and naive-rollouts into a single group so that the relative-advantage is dominated by hint-rollouts, with the model learning to directly imitate the off-policy hint distribution rather than exploring with its own policy under hint guidance.
Specifically, externally guided hint-rollouts are typically simpler than naive-rollouts, leading to more trajectories with positive advantages that dominate the update signal and bias the policy toward fitting the off-policy distribution. 
Additionally, hints from external reasoning models are usually much longer than the policy's own outputs, further exacerbating this imbalance.
As shown in Figure~\ref{fig1:sub_b}, the average reward of hint-rollouts increases steadily, whereas that of naive-rollouts collapses, causing the model (denoted as Baseline) to eventually lose the ability to perform inference without hints.

To address these issues, we propose \textbf{ADHint} (\textbf{AD}aptive \textbf{Hint}s with Difficulty Priors for Reinforcement Learning), which incorporates difficulty into both the hint-ratio schedule and the relative-advantage estimation to achieve a principled trade-off between exploration and imitation. 
For the hint-ratio schedule, we propose \textbf{Adaptive Hint with Sample Difficulty Prior} (AH-SDP), which adaptively determines each sample's hint ratio from a difficulty prior estimated from the average reward of its naive-rollouts, maintaining the resulting hint-rollouts within a moderate difficulty range to provide stable update signals.
Furthermore, to prevent drastic shifts toward the off-policy distribution that trigger entropy collapse, we propose \textbf{Consistency-based Gradient Modulation} (CGM), which downweights the gradients of hint tokens whose entropy deviates significantly from that of the policy-generated continuation.
We also introduce \textbf{Selective Masking for Hint Preservation} (Selective Masking) to discard erroneous update signals from hint-rollouts with negative advantages.
For the relative-advantage estimation, we propose \textbf{Advantage Estimation with Rollout Difficulty Posterior} (\textsc{AE-RDP}), which constructs the rollout difficulty posterior from the average rewards of both naive-rollouts and hint-rollouts, and uses it to estimate advantages. Positive naive-rollouts with greater difficulty receive larger advantages, as they provide valuable learning signals better aligned with the current policy, whereas negative hint-rollouts with lower difficulty are penalized more heavily.

We conduct extensive experiments across multiple modalities, model scales, model families, domains, and tasks. As shown in Figure~\ref{fig:2}, ADHint consistently yields substantial gains, demonstrating synergistic improvements in both reasoning ability and out-of-distribution generalization. Our main contributions are summarized as follows:
\begin{itemize}
    \item We reveal that difficulty is a crucial signal for both the hint-ratio schedule and relative-advantage estimation, and that neglecting it results in unstable learning and overfitting to the off-policy distribution.

    \item We propose ADHint, which leverages sample difficulty priors and rollout difficulty posteriors for the hint-ratio schedule and relative-advantage estimation, leading to a better balance between exploration and imitation.

    \item Experiments across diverse settings demonstrate ADHint's superiority, delivering robust and significant performance gains across comprehensive multi-domain benchmarks.
\end{itemize}
\section{Related Work}
\label{sec:related}
\subsection{Multimodal Reinforcement Learning}
GRPO~\cite{shao2024deepseekmath} replaces PPO's critic~\cite{schulman2017proximal} with group-relative advantage estimation and underpins recent reasoning models~\cite{team2025kimi,tan2025towards}. 
In parallel, multimodal RL studies have explored reasoning styles~\cite{wang2025vl,zheng2025deepeyes}, rollout strategies~\cite{liu2025noisyrollout,wang2025perception}, reward design~\cite{liu2025visionreasoner,liu2025visual}, and data augmentation~\cite{liang2025modomodomultidomaindatamixtures}. 
Nevertheless, RL algorithms mainly unlock the latent potential of pretrained models and often fail to expand their fundamental capability boundary.
\subsection{Reinforcement Learning with Hints}
Hint-based RL injects off-policy solution prefixes to guide reasoning while retaining on-policy exploration.
Existing methods anneal hint ratios over time~\cite{liu2025uft,huang2025blending}, mix rollouts with different hint levels within each group~\cite{zhang2025stephint}, or learn instance-specific ratios through multi-round training~\cite{li2025staying}.
Others restrict hints to the most challenging  samples and select the ratio via uniform sampling~\cite{chen2025data}, binary search~\cite{zhang2025bread}, or progressive search~\cite{Huang_2025_ICCV,liu2025ghpo}.
In contrast, ADHint explicitly incorporates difficulty into both the hint-ratio schedule and the relative-advantage estimation to address the unstable learning and excessive imitation observed in existing hint-based methods.
Detailed discussion is provided in Appendix~\ref{Appendix:C}.
\begin{figure*}[!t]
    \centering
    \captionsetup{skip=4pt}
    \includegraphics[width=1.0\linewidth]{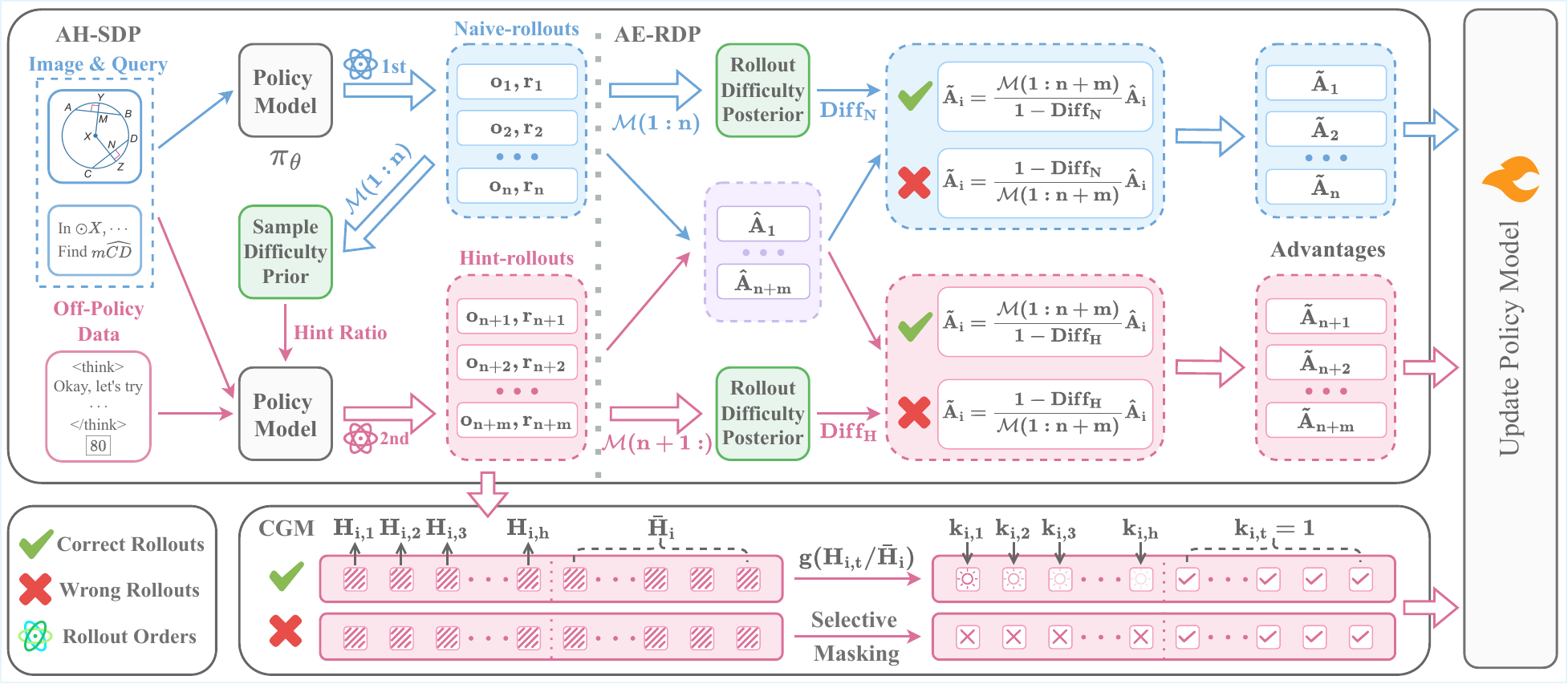}
    \caption{\textbf{An overview of ADHint.} For each sample, ADHint successively infers twice to obtain naive-rollouts and hint-rollouts, where the hint ratio of the latter is scheduled based on the difficulty evaluated from the former. It then leverages the relative difficulty of naive-rollouts and hint-rollouts to estimate their respective advantages, and modulates the gradient of each hint token based on the consistency of entropy.}
    \label{fig:overview}
    \vspace{-1.2em}
\end{figure*}
\section{Preliminary}
\label{sec:preliminary}
\subsection{Group Relative Policy Optimization}
For each query $q$ in the question set $Q$, GRPO samples a group of rollouts $\{o_1, \dots, o_G\}$ from the old policy model 
$\pi_{\theta_{\text{old}}}$, and computes their rewards 
$\{r_1, \dots, r_G\}$.
The current policy model $\pi_{\theta}$ is optimized by maximizing 
the following objective:\par\vspace{-0.8em}
{\scriptsize
\begin{equation*}
\label{equ:1}
\begin{aligned}
&\mathcal{J}_{\text{GRPO}}(\pi_\theta)
= \mathbb{E}_{q \sim Q,\,\{o_i\}_{i=1}^G \sim
    \pi_{\theta_{\text{old}}}(\cdot \mid q)} \\[-1pt]
&\frac{1}{G} \sum_{i=1}^G
    \frac{1}{|o_i|} \sum_{t=1}^{|o_i|}
    \Biggl\{
    \frac{\pi_{\theta}(o_{i,t} \mid q, o_{i,<t})}
         {\pi_{\theta_{\text{old}}}(o_{i,t} \mid q, o_{i,<t})}
    \hat{A}_i
    - \beta \, D_{\mathrm{KL}}
    \bigl(\pi_\theta \,\|\, \pi_{\text{ref}}\bigr)
    \Biggr\}.
\end{aligned}
\end{equation*}
}
where the standard \emph{min} and \emph{clip} operations for avoiding extreme gradients are omitted for clarity, and $D_{\text{KL}}(\pi_\theta \| \pi_{\text{ref}})$ is a KL constraint term. $\hat{A}_i$ is the relative advantage, calculated by normalizing the reward within the group:
\[
\hat{A}_i = \frac{r_i - \text{mean}(\{r_1, \dots, r_G\})}{\text{std}(\{r_1, \dots, r_G\})}. 
\]
The gradient for a given $q$ and $o_i$ is calculated as:\par\vspace{-1.0em}
{\small
\begin{equation*}
\label{equ:grad}
\begin{aligned}
&\nabla_{\theta} \mathcal{J}^*_{\text{GRPO}}(\pi_{\theta})
  \Big|_{q,o_i}
  = \frac{1}{G \cdot |o_i|} \sum_{t=1}^{|o_i|} \\[-1pt]
&\quad
   \frac{\pi_{\theta}(o_{i,t} \mid q, o_{i,<t})}
        {\pi_{\theta_{\text{old}}}(o_{i,t} \mid q, o_{i,<t})}
   \cdot \hat{A}_i
   \cdot \nabla_{\theta}
   \log\!\bigl(\pi_{\theta}(o_{i,t} \mid q, o_{i,<t})\bigr).
\end{aligned}
\end{equation*}
}
\subsection{Hint-based On-Policy Learning}
Recent works use the initial part of off-policy reasoning trajectories from a teacher model as \emph{hints} to guide rollouts. 
Given a query $q$, a hint ratio $w$ specifies the prefix length $h$ copied from the off-policy trajectory. 
Conditioned on $q$, the policy model $\pi_\theta$ then produces two types of rollouts:
\begin{itemize}[leftmargin=*]
    \item \textbf{Naive-rollouts:} Sample $n$ unguided rollouts $\{o_1,\dots,o_n\}$, where $o_i \sim \pi_{\theta}(\cdot \mid q)$.
    
    \item \textbf{Hint-rollouts:} Sample $m$ hint-guided rollouts $\{o_{n+1},\dots,o_{n+m}\}$, each constructed as
    $o_i=[\,o_{i,<h};o_{i,>h}\,]$, where $o_{i,<h}$ is the hint prefix
    of length $h$, and $o_{i,>h}$ is the continuation sampled as
    $o_{i,>h}\sim\pi_{\theta}(\cdot\mid q,o_{i,<h})$.
\end{itemize}

In standard practice, the $n{+}m$ rollouts are pooled into a single group to compute relative advantages.
Notably, since hint tokens ($t \le h$) are not generated by the policy model, following prior work~\cite{yan2025learning}, we set $\pi_{\theta_{\text{old}}}(o_{i,t} \mid q, o_{i,<t}) = 1$ for all hint tokens in hint-rollouts.
\section{Method}
\label{sec:method}
\vspace{-2mm}
\subsection{Overview}
As shown in Figure~\ref{fig:overview}, ADHint consists of four modules: 
(1) \textbf{Adaptive Hint with Sample Difficulty Prior} (AH-SDP), which sequentially generates naive- and hint-rollouts, using the sample difficulty prior evaluated from the former to schedule the hint ratio for the latter; 
(2) \textbf{Advantage Estimation with Rollout Difficulty Posterior} (AE-RDP), which uses the rollout difficulty posterior to estimate relative advantages across correct and incorrect naive- and hint-rollouts;
(3) \textbf{Consistency-based Gradient Modulation} (CGM), which modulates each hint token's gradient based on the consistency between its entropy distribution and that of the continuation; and 
(4) \textbf{Selective Masking for Hint Preservation} (Selective Masking), which masks hint-token updates in hint-rollouts that remain incorrect despite guidance.
Additionally, Algorithm~\ref{algo} summarizes the overall procedure. 
\subsection{Adaptive Hint with Sample Difficulty Prior}
Assigning the same hint ratio to every sample in the current update batch~\cite{zhang2025stephint,huang2025blending,liu2025uft} introduces a mismatch in the difficulty of hint-rollouts across samples, yielding unstable training signals and reducing learning efficiency.
As shown in Figure~\ref{fig1:sub_a}, entropy drops rapidly early in training but rises sharply as the hint ratio decreases, suggesting that the model memorizes superficial hint patterns rather than internalizing the underlying knowledge.

Unlike existing methods, we schedule each sample's hint ratio using the difficulty prior evaluated from its naive-rollouts.
This keeps hint-rollouts within a moderate-difficulty regime (Figure~\ref{fig5:sub_a}) and provides low-variance update signals at each step.
For each query $q$, we define the difficulty score of naive-rollouts as\par\vspace{-0.6em}
{\small
\begin{equation}
\text{Diff}_\text{N} \;=\; 1 - \mathcal{M}(1:n)
\;=\; 1 - \operatorname{mean}(r_1, \dots, r_n) .
\label{eq:diff_naive}
\end{equation}
}
We then map this score to the hint ratio $w$ using a linear function:\par
\begin{figure}[!t]
	\centering
	\captionsetup{skip=4pt}
	\captionsetup[subfigure]{skip=2pt,justification=centering}
	\begin{subfigure}[t]{0.49\linewidth}
		\centering
		\includegraphics[width=\linewidth]{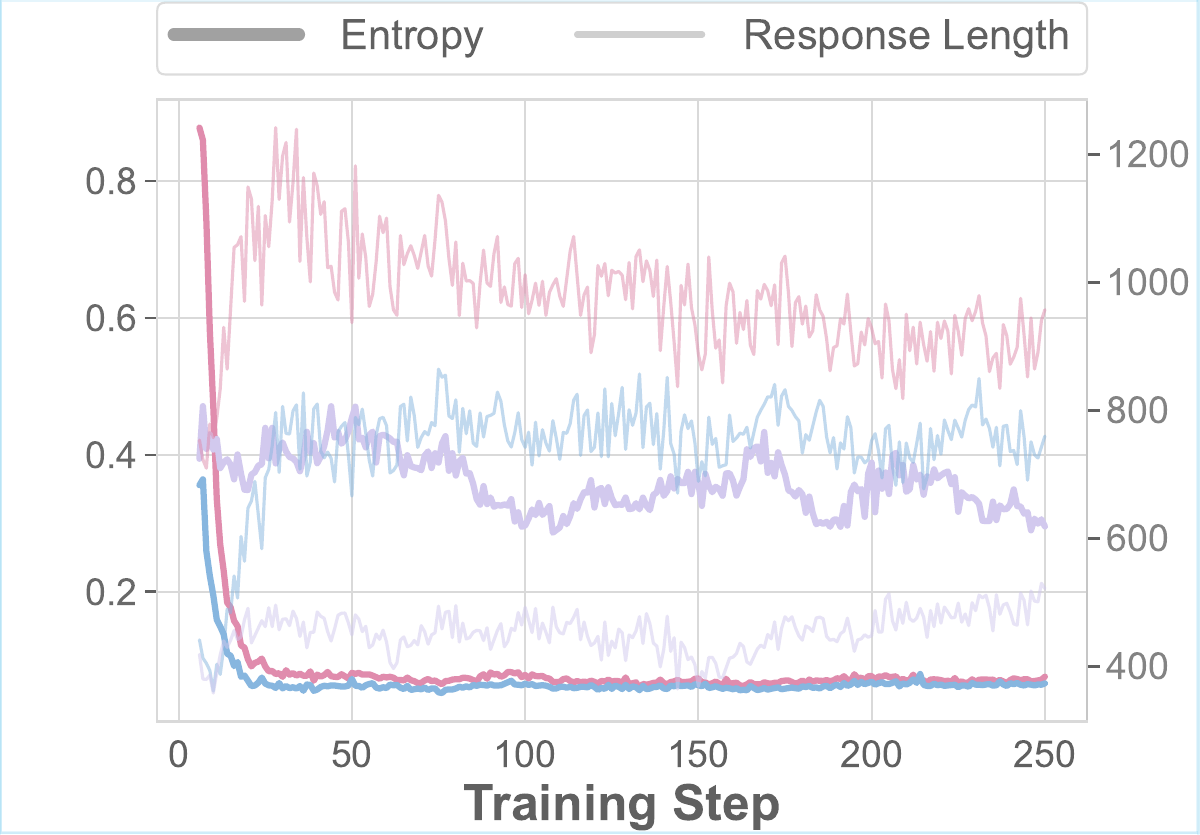}
		\caption{ADHint w/o\protect\\CGM} 
		\label{fig4:sub_a}
	\end{subfigure}
	\hfill
	\begin{subfigure}[t]{0.49\linewidth}
		\centering
		\includegraphics[width=\linewidth]{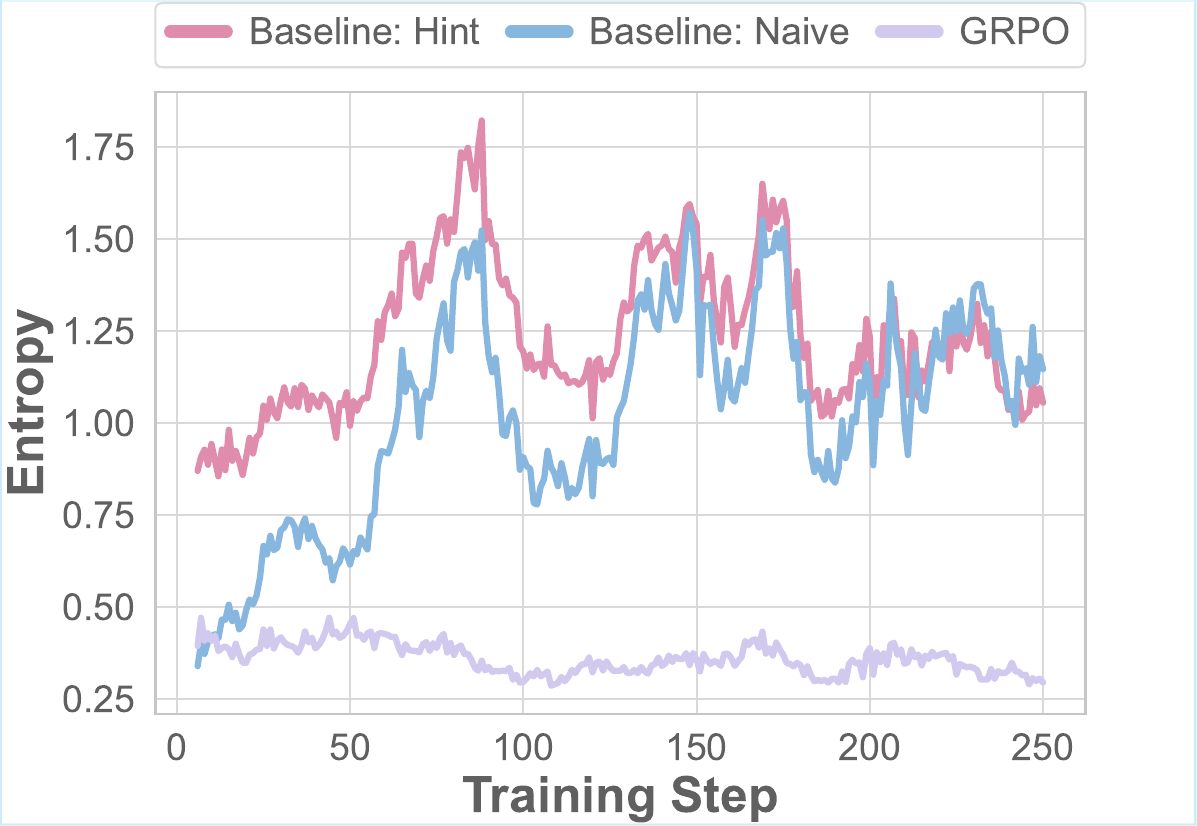}
		\caption{ADHint w/o\protect\\Selective Masking} 
		\label{fig4:sub_b}
	\end{subfigure}
	\caption{
		\textbf{Baseline behavior after removing CGM or Selective Masking.}
		The left subfigure shows the baseline's entropy (left axis) and response length (right axis).
		The right subfigure compares the entropy of the baseline and GRPO, where ``Baseline: Hint/Naive'' is computed on hint- or naive-rollouts only.
	}
	\label{fig4}
	\vspace{-1.em}
\end{figure}

\vspace{-0.6em}
{\small
\begin{equation}
w = f(\text{Diff}_\text{N})
= (w_{\text{max}} - w_{\text{min}})\cdot \text{Diff}_\text{N}
+ w_{\text{min}} + \sigma .
\label{eq:hint_ratio}
\end{equation}
}
where $w_{\text{max}}$ and $w_{\text{min}}$ are the maximum and minimum ratios, and $\sigma \sim \mathcal{U}(-R, R)$ is uniformly distributed noise. We fix $w_{\text{min}} = 0$ since easy questions require no hints, thus eliminating it as a hyperparameter. The resulting hint ratio determines the hint length $h$ used to generate $m$ hint-rollouts $\{o_{n+1},\dots,o_{n+m}\}$.
\paragraph{Consistency-based Gradient Modulation.}
Off-policy hints may differ substantially from the on-policy model in language style, knowledge structure, and response length, making indiscriminate learning prone to overfitting.
As shown in Figure~\ref{fig4:sub_a}, the entropies of both naive- and hint-rollouts collapse sharply, while naive-rollout lengths rapidly converge to those of hint-rollouts, indicating policy drift toward the off-policy hints.
To mitigate this, we measure each hint token's consistency with the generated continuation, which reflects the model's intrinsic distribution, and modulate its gradient accordingly to prevent destructive distribution shifts.
Specifically, the entropy of each hint token in a hint-rollout $o_i$ is defined as
\begin{equation}
H_{i,t} = \mathcal{H}(\pi_{\theta}(o_{i,t} \mid q, o_{i,<t})), 
\label{eq:token_entropy}
\end{equation}
and the average entropy of the policy-generated continuation as
\begin{equation}
\bar{H}_i = \frac{1}{|o_{i,>h}|} \sum_{t=h+1}^{|o_i|} \mathcal{H}(\pi_{\theta}(o_{i,t} \mid q, o_{i,<t})). 
\label{eq:seq_entropy}
\end{equation}
We then scale the gradient of each token by
\begin{equation*}
k_{i,t} =
\begin{cases}
g(H_{i,t} / \bar{H}_i;\alpha), & \text{if } 1 \le t \le h, \\
1, & \text{otherwise.}
\end{cases}
\end{equation*}
where $g(x;\alpha)$ is a symmetric cosine-based schedule centered at $g(1;\alpha)=1$, with a ramp-up on $[\alpha, 1]$ and a ramp-down on $[1, 1/\alpha]$ for $0<\alpha\le 1$:
\[
g(x;\alpha) =
\begin{cases}
\sin\bigl(\frac{\pi}{2} \cdot \frac{x - \alpha}{1 - \alpha}\bigr), & \alpha \le x \le 1, \\[4pt]
\cos\bigl(\frac{\pi}{2} \cdot \frac{x - 1}{1/\alpha - 1}\bigr), & 1 < x \le 1/\alpha, \\[4pt]
0, & \text{otherwise}.
\end{cases}
\]
\paragraph{Selective Masking for Hint Preservation.}
Adaptive hinting keeps hint-rollouts moderately difficult, leaving some trajectories with negative advantages.
Applying negative updates to the hint prefix $o_{i,\le h}$, which is assumed to be correct, introduces conflicting gradients, destabilizes training, and often triggers entropy spikes by increasing the model's predictive uncertainty (Figure~\ref{fig4:sub_b}).
We therefore selectively mask hint-token gradients:
\begin{equation}
k_{i,t} =
\begin{cases}
0, & 1 \le t \le h \text{ and } \hat{A}_i \le 0, \\
g(H_{i,t} / \bar{H}_i;\alpha), & 1 \le t \le h \text{ and } \hat{A}_i > 0, \\
1, & t > h.
\end{cases}
\label{eq:factor}
\end{equation}
For notational consistency, we set $k_{i,t}=1$ for naive-rollouts.
\begin{table*}[t!]
	\centering
	\captionsetup{skip=4pt}
	\renewcommand{\arraystretch}{1.0}
	\footnotesize
	\setlength{\tabcolsep}{4.0pt}
	\begin{tabular}{@{}l ccc cc c c@{}}
		\toprule[1pt]
		\multirow{2}{*}{\textbf{Method}} & \multicolumn{3}{c}{\textbf{Math Reasoning}} & \multicolumn{2}{c}{\textbf{Multi-Discipline}} & \textbf{Logical} & \multirow{2}{*}{\textbf{Avg.}} \\
		\cmidrule(lr){2-4} \cmidrule(lr){5-6} \cmidrule(lr){7-7}
		& MathVista & MathVerse & We-Math & MMMU & MMMU-Pro & LogicVista & \\
		\midrule[0.8pt]
		
		\rowcolor{gray!10}
		\multicolumn{8}{c}{\textit{Base MLLM: Qwen2.5-VL-3B}} \\
		
		SFT &  51.9/61.3& 31.2/\textbf{54.2} &32.5/58.5  &28.6/46.3  &17.1/29.4  &18.1/37.8  &29.9/47.9 \\ 
		SFT $\rightarrow$ RL & 53.6/\underline{63.3} &37.4/\underline{54.1}  &38.3/\textbf{60.6}  &30.6/47.7  &18.7/\underline{30.9}  &18.1/\underline{39.9}  &32.8/\underline{49.4}  \\
		GRPO~\cite{shao2024deepseekmath} & \underline{63.9}/63.1 & \underline{43.1}/45.0 & \underline{59.0}/59.0 & \underline{47.5}/\underline{49.1} & \underline{30.7}/30.1 & \textbf{43.8}/38.9 & \underline{48.0}/47.5  \\ 
		
		\midrule
		LUFFY~\cite{yan2025learning} & 41.2/45.6 & 31.7/35.0  & 42.3/46.1  &34.3/44.6  &20.3/23.1  &35.4/38.3  &34.2/38.8  \\ 
		StepHint~\cite{zhang2025stephint} &34.9/44.6  &28.7/32.5  &32.0/40.2  &18.3/35.7  &19.8/26.0  &12.5/27.8  & 24.4/34.5  \\ 
		HintGRPO~\cite{Huang_2025_ICCV} &  56.7/52.5 & 37.4/39.7 &42.0/40.5  &44.4/43.0  &27.5/25.1  &35.7/33.7  &40.6/39.1  \\ 
		GHPO~\cite{liu2025ghpo} & 57.3/55.2 & 35.6/35.4 & 41.0/39.1 &44.4/42.6  &27.0/26.3  &35.3/33.9  & 40.1/38.8 \\ 
		
		\midrule[0.8pt]
		\rowcolor{green!15} 
		\textbf{ADHint (Ours)} & \textbf{64.8}/\textbf{64.3} & \textbf{49.5}/50.6 & \textbf{60.4}/\underline{59.1} & \textbf{51.4}/\textbf{50.0} & \textbf{33.3}/\textbf{32.0} & \underline{42.6}/\textbf{41.4} & \textbf{50.3}/\textbf{49.6} \\ 
		
		\midrule[0.8pt]
		\rowcolor{gray!10}
		\multicolumn{8}{c}{\textit{Base MLLM: Qwen2.5-VL-7B}} \\
		
		SFT & 59.0/68.7 & 40.1/\underline{64.0}  & 44.1/70.1 & 31.2/51.0 & 20.6/35.6 & 23.7/46.0 & 36.5/55.9 \\
		SFT $\rightarrow$ RL & 64.7/71.1 & \underline{56.4}/\textbf{67.2} & 54.1/\textbf{72.5} & 35.9/54.2 & 25.8/38.3 & 28.8/\underline{47.8} & 44.3/\underline{58.5}  \\
		GRPO~\cite{shao2024deepseekmath} & \underline{73.4}/\underline{73.4}  & 55.3/59.2  & \underline{69.8}/68.8 & \underline{55.3}/\underline{55.6} & \underline{40.4}/\underline{40.0} & \underline{44.9}/46.4 & \underline{56.5}/57.2 \\ 
		
		\midrule
		LUFFY~\cite{yan2025learning} &63.6/61.2 &34.6/38.1 &52.2/54.8 &44.6/47.4 &30.6/29.7 &40.2/39.2 &44.3/45.1 \\ 
		StepHint~\cite{zhang2025stephint} &47.2/49.5 &27.7/31.4 &39.8/41.1 &18.6/27.9  &9.5/19.3& 10.5/22.7& 25.6/32.0 \\ 
		HintGRPO~\cite{Huang_2025_ICCV} & 65.9/64.3 & 45.9/47.5 & 42.9/44.7 & 40.6/39.3 & 30.2/28.7 & 36.4/34.7 & 43.7/43.2 \\ 
		GHPO~\cite{liu2025ghpo} & 66.1/63.9 & 52.3/48.9 & 63.0/52.8 & 51.7/50.0 & 35.3/32.3 & 42.4/40.5 & 51.8/48.1\\ 
		
		\midrule[0.8pt]
		\rowcolor{green!15} 
		\textbf{ADHint (Ours)} & \textbf{74.4}/\textbf{74.8} & \textbf{60.6}/61.8 & \textbf{69.9}/\underline{70.8} & \textbf{56.4}/\textbf{56.4} & \textbf{41.3}/\textbf{41.8} & \textbf{48.7}/\textbf{49.4} & \textbf{58.6}/\textbf{59.2} \\ 
		
		\bottomrule[1pt]
	\end{tabular}
	\caption{
		\textbf{Comparative performance on Qwen2.5-VL models.}
		For each benchmark, we report \textit{pass@1} (the first value) and \textit{avg@8} (the second value).
		The upper block uses Qwen2.5-VL-3B as the base MLLM, while the lower block uses Qwen2.5-VL-7B.
		The best and runner-up results are highlighted in \textbf{bold} and \underline{underlined}, respectively.
	}
	\label{tab:1}
	\vspace{-1.5em}
\end{table*}
\subsection{Advantage Estimation with Rollout Difficulty Posterior}
Pooling naive- and hint-rollouts for relative-advantage estimation~\cite{zhang2025stephint,huang2025blending} biases updates toward hint-rollouts, which typically contain more correct trajectories and longer responses. As shown in Figure~\ref{fig1:sub_b}, hint-rollout rewards continue to rise while naive-rollout rewards collapse, indicating overfitting to the hint distribution.
Consequently, the policy improves ``text-completion'' behavior but sacrifices genuine reasoning ability. To correct this bias, we introduce the rollout difficulty posterior for advantage estimation, which characterizes the relative difficulty between naive-rollouts and hint-rollouts after all rollouts are completed. 

Our design follows two intuitions. Positive naive-rollouts provide more informative update signals because they are harder and fully generated by the current policy, whereas negative hint-rollouts warrant stronger penalties because they are easier.
Accordingly, for each query $q$, we define the difficulty score of naive-rollouts in Eq.~\ref{eq:diff_naive} and analogously define that of hint-rollouts as \par\vspace{-0.6em}
{\small
	\begin{equation}
		\text{Diff}_\text{H} \;=\; 1 - \mathcal{M}(n+1:)
		\;=\; 1 - \operatorname{mean}(r_{n+1}, \dots, r_{n+m}) .
		\label{eq:diff_hint}
	\end{equation}
}
Then the difficulty score of rollout $o_i$ is denoted as
\begin{equation*}
\mathrm{Diff}_i =
\begin{cases}
\mathrm{Diff}_\mathrm{N}, & \text{if } 1 \le i \le n, \\
\mathrm{Diff}_\mathrm{H}, & \text{if } n+1 \le i \le n+m.
\end{cases}
\label{eq:diff_i}
\end{equation*}
Let $\hat{A}_i$ denote the relative advantage of rollout $o_i$, estimated by pooling all $n+m$ trajectories. We define a sign indicator $s_i = \operatorname{sgn}(\hat{A}_i)$, where $s_i = -1$ when $\hat{A}_i \le 0$. Finally, the relative advantages in ADHint are estimated as
\begin{equation}
\tilde{A}_i = \left( \frac{\mathcal{M}(1:n+m)}{1 - \mathrm{Diff}_i} \right)^{s_i} \hat{A}_i .
\label{eq:adv_rdp}
\end{equation}
\subsection{Overall Gradient Computation}
Building on the above mechanisms, we compute the gradient for a given query $q$ and rollout $o_i$ as:\par\vspace{-1.0em}
{\small
\begin{equation}
\label{eq:final_grad}
\begin{aligned}
&\nabla_{\theta} \mathcal{J}^*_{\text{ADHint}}(\pi_{\theta})
  \Big|_{q,o_i}
  = \frac{1}{G \cdot |o_i|} \sum_{t=1}^{|o_i|} k_{i,t} \\[-1pt]
&\quad
   \frac{\pi_{\theta}(o_{i,t} \mid q, o_{i,<t})}
        {\pi_{\theta_{\text{old}}}(o_{i,t} \mid q, o_{i,<t})}
   \cdot \tilde{A}_i \cdot
   \nabla_{\theta}
   \log\!\bigl(\pi_{\theta}(o_{i,t} \mid q, o_{i,<t})\bigr).
\end{aligned}
\end{equation}
}

Notably, to avoid inaccurate difficulty prior evaluations caused by formatting errors, we warm up the policy model with GRPO for the first five steps to ensure the model follows the required response format (Figure~\ref{fig5:sub_f}).
\begin{table*}[t!]
	\centering
	\captionsetup{skip=4pt}
	\renewcommand{\arraystretch}{1.0}
	\footnotesize
	\setlength{\tabcolsep}{4.0pt}
	\begin{tabular}{@{}lccccccc@{}}
		\toprule[1pt]
		\textbf{Method} & MathVista & MathVerse & We-Math & MMMU & MMMU-Pro & LogicVista & \textbf{Avg.} \\
		\midrule[0.8pt]
		
		\rowcolor{gray!10}
		\multicolumn{8}{c}{\textit{Base MLLM: Qwen3-VL-4B}} \\
		
		GRPO~\cite{shao2024deepseekmath} & \textbf{73.6}/\textbf{77.9} & 56.1/\underline{75.9} & 67.9/\textbf{82.1} & 38.1/49.5 & 39.4/\underline{51.2} & 46.2/\underline{60.8} & 53.6/\underline{66.2} \\ 
		LUFFY~\cite{yan2025learning} & 64.7/72.3 & 56.3/70.9 & 56.6/74.5 & 29.9/46.6 & 32.0/49.7 & 38.2/57.1 & 46.3/61.9 \\ 
		StepHint~\cite{zhang2025stephint} & 66.2/71.2 & 53.7/70.2 & 58.2/74.0 & 34.7/\underline{52.5} & 32.8/49.4 & 38.0/57.0 & 47.3/62.4 \\ 
		HintGRPO~\cite{Huang_2025_ICCV} & 67.1/71.9 & \underline{60.3}/72.8 & 64.6/77.6 & \underline{42.0}/52.2 & \underline{40.1}/49.4 & 48.0/57.2 & 53.7/63.5 \\ 
		GHPO~\cite{liu2025ghpo} & 70.4/75.9 & 54.9/70.5 & \textbf{69.5}/76.4 & 41.2/48.5 & 39.8/51.1 & \underline{48.4}/59.0 & \underline{54.0}/63.6 \\
		
		\midrule[0.8pt]
		\rowcolor{green!15} 
		\textbf{ADHint (Ours)} & \underline{71.7}/\underline{77.6} & \textbf{63.3}/\textbf{76.1} & \underline{69.3}/\underline{81.0} & \textbf{50.3}/\textbf{65.6} & \textbf{43.2}/\textbf{52.3} & \textbf{50.6}/\textbf{62.4} & \textbf{58.1}/\textbf{69.2} \\
		
		\midrule[0.8pt]
		\rowcolor{gray!10}
		\multicolumn{8}{c}{\textit{Base MLLM: Qwen3-VL-8B}} \\
		
		GRPO~\cite{shao2024deepseekmath} & \textbf{74.6}/\underline{78.8} & 63.1/\underline{79.0} & \underline{73.1}/\underline{82.6} & \underline{61.0}/\textbf{71.7} & \underline{46.5}/\textbf{57.9} & \underline{49.4}/\underline{63.0} & \underline{61.3}/\underline{72.2} \\ 
		LUFFY~\cite{yan2025learning} & 54.4/53.7 & 28.0/18.6 & 26.0/18.1 & 21.0/26.6 & 21.2/27.2 & 14.5/21.8 & 27.5/27.7 \\ 
		StepHint~\cite{zhang2025stephint} & 55.0/54.8 & 28.4/20.5 & 25.8/17.5 & 22.5/27.7 & 22.3/27.2 & 20.5/22.5 & 29.1/28.4 \\ 
		HintGRPO~\cite{Huang_2025_ICCV} & 64.9/64.4 & \underline{64.2}/68.7 & 68.6/71.7 & 47.2/55.5 & 38.4/\underline{46.0} & 43.1/51.9 & 54.4/59.7 \\ 
		GHPO~\cite{liu2025ghpo} & 64.9/63.2 & 39.7/37.7 & 49.9/48.8 & 38.8/38.5 & 40.4/40.0 & 45.1/43.6 & 46.5/45.3 \\
		
		\midrule[0.8pt]
		\rowcolor{green!15} 
		\textbf{ADHint (Ours)} & \underline{74.0}/\textbf{79.6} & \textbf{75.2}/\textbf{80.1} & \textbf{76.4}/\textbf{85.6} & \textbf{64.4}/\underline{70.1} & \textbf{51.8}/\textbf{57.9} & \textbf{56.3}/\textbf{64.9} & \textbf{66.4}/\textbf{73.0} \\
		
		\bottomrule[1pt]
	\end{tabular}
	\caption{
		\textbf{Comparative performance on Qwen3-VL models.}
	}
	\label{tab:2}
	\vspace{-1.em}
\end{table*}
\begin{table*}[t!]
	\centering
	\captionsetup{skip=4pt}
	\renewcommand{\arraystretch}{0.85}
	\footnotesize
	\setlength{\tabcolsep}{6.0pt}
	\begin{tabular}{@{}lccccccc@{}}
		\toprule[1pt]
		\textbf{Method} 
		& MathVista & MathVerse & We-Math & MMMU & MMMU-Pro & LogicVista & \textbf{Avg.} \\
		\midrule[0.8pt]
		
		\rowcolor{green!15}
		\textbf{ADHint} 
		& 74.4/\textbf{74.8} 
		& 60.6/\textbf{61.8} 
		& \textbf{69.9}/\textbf{70.8} 
		& \textbf{56.4}/\textbf{56.4} 
		& 41.3/\textbf{41.8} 
		& \textbf{48.7}/\textbf{49.4} 
		& \textbf{58.6}/\textbf{59.2} \\
		
		\midrule[0.4pt]
		
		w/o AH-SDP 
		& 74.1/73.5 
		& 58.7/58.9 
		& 68.6/67.4 
		& 54.6/55.0 
		& 41.3/40.3 
		& 48.2/48.4 
		& 57.6/57.3 \\
		
		w/o AE-RDP 
		& 74.3/74.1 
		& \textbf{61.5}/60.0 
		& 67.5/68.4 
		& 55.9/56.1 
		& 41.2/41.1 
		& 45.8/46.3 
		& 57.7/57.7 \\
		
		w/o CGM 
		& 73.1/72.8 
		& 53.9/55.0 
		& 68.2/68.0 
		& 56.3/51.1 
		& 40.0/40.0 
		& 50.5/48.2 
		& 57.0/55.9 \\
		
		w/o Selective Masking 
		& \textbf{74.7}/73.7 
		& 58.5/61.0 
		& 67.4/67.6 
		& 55.8/55.9 
		& \textbf{41.6}/40.9 
		& 47.3/46.3 
		& 57.6/57.6 \\
		
		\bottomrule[1pt]
	\end{tabular}
	\caption{
		\textbf{Ablation study of ADHint.}
	}
	\label{tab:6}
	\vspace{-1.2em}
\end{table*}
\section{Experiments}
\label{sec:experiments}
\subsection{Settings}
We evaluate ADHint across modalities, model scales, model families, and domains.
For general MLLM reasoning (e.g., math and chart reasoning), we use Qwen2.5-VL-7B-Instruct~\cite{bai2025qwen2}, Qwen3-VL-8B-Instruct~\cite{bai2025qwen3vltechnicalreport}, Qwen2.5-VL-3B-Instruct, Qwen3-VL-4B-Instruct, and MiMo-VL-7B-SFT-2508~\cite{coreteam2025mimovltechnicalreport}.
We assess cross-domain generalization on Medical VQA with Qwen2.5-VL-7B-Instruct. For LLM math reasoning, we use Qwen2.5-Math-7B~\cite{yang2024qwen2}, following the data and experimental settings of GHPO~\cite{liu2025ghpo}.
\paragraph{Training data.}
For MLLMs, we construct the training data from ViRL39K~\cite{wang2025vl}, which spans STEM, chart, document, and spatial reasoning.
We sample hints using Qwen3-VL-30B-A3B-Thinking and Qwen3-VL-235B-A22B-Thinking, and remove examples that are excessively simple for Qwen2.5-VL-7B-Instruct, yielding 17K training queries termed ViRL-Hint17k.
For Medical VQA, we use a 15K subset of the PMC-VQA training set~\cite{zhang2023pmc}. 
For LLMs, we use NuminaMath-S, a mixture of MATH~\cite{hendrycks2measuring} and NuminaMath-1.5~\cite{li2024numinamath} curated by GHPO. 
Details are provided in Appendix~\ref{Appendix:D.1}.
\paragraph{Benchmarks.}
For MLLMs, we evaluate math reasoning on MathVista~\cite{lu2023mathvista}, MathVerse~\cite{zhang2024mathverse}, and We-Math~\cite{qiao2024we}; multi-discipline reasoning on MMMU~\cite{yue2024mmmu} and MMMU-Pro~\cite{yue2024mmmupro}; and logical reasoning on LogicVista~\cite{xiao2024logicvista}.
For Medical VQA, we use the PMC-VQA test set (PMC-Test)~\cite{zhang2023pmc}. For LLMs, we evaluate math reasoning on AIME~\cite{patel2024aime}, AMC~\cite{li2024numinamath}, Minerva~\cite{lewkowycz2022solving}, OlympiadBench~\cite{he2024olympiadbench}, and MATH500.
Details are provided in Appendix~\ref{Appendix:D.2}.
\paragraph{Evaluation Metrics.}
For MLLMs, we report \textit{pass@1} and \textit{avg@8} for each benchmark~\cite{wang2025vl,wang2025perception}. \textit{Pass@1} measures the accuracy of a single rollout per sample under greedy decoding, reflecting the model's out-of-distribution reasoning generalization. \textit{Avg@8} computes the average accuracy over eight rollouts per sample at temperature 1.0, reflecting the model's robustness and mastery of the underlying knowledge. 
For Medical VQA, we report \textit{pass@1}~\cite{zhu2025toward}. 
For LLMs, we report either \textit{pass@1} or \textit{avg@32} depending on the benchmark to mitigate variance introduced by the small sizes of some datasets~\cite{liu2025ghpo}. 
\paragraph{Implementation Details.}
For MLLMs, we train for 250 steps (about 2 epochs) with a learning rate of $1 \times 10^{-6}$. We set the rollout batch size to 128 and the maximum generation length to 8K. 
For LLMs, we train for 5 epochs with an initial learning rate of $1 \times 10^{-6}$, decayed to zero using a cosine schedule with a warm-up phase covering 10\% of the total global steps. We set the maximum generation length to 2K. We set $w_{\text{max}}$ to 0.2 for MLLMs and 0.4 for LLMs, and use $\alpha = 0.5$ as the CGM threshold for both. All training is performed across four compute nodes. Further details are provided in Appendix~\ref{Appendix:D.3}.
\paragraph{Baselines.}
We compare ADHint against seven baselines:
(1) \textbf{GRPO};
(2) \textbf{SFT}, trained for two epochs on the full training set;
(3) \textbf{SFT $\rightarrow$ RL}, with two SFT epochs followed by GRPO;
(4) \textbf{LUFFY}~\cite{yan2025learning}, which jointly performs on-policy and off-policy updates at each step;
(5) \textbf{StepHint}~\cite{zhang2025stephint}, which groups hint-rollouts at multiple ratios with naive-rollouts and ground-truth trajectories;
(6) \textbf{HintGRPO}~\cite{Huang_2025_ICCV}, which progressively increases the hint ratio for the hardest samples to obtain positive trajectories; and
(7) \textbf{GHPO}~\cite{liu2025ghpo}, which treats hints as part of the query and applies multiple hint ratios to the hardest samples.
We note that some baselines crash during training, and we report results from their last checkpoint available before the crash. 
More detailed information about the baselines and their reimplementation is provided in Appendices~\ref{Appendix:D.4} and~\ref{Appendix:D.5}, respectively.
\subsection{Main Results}
\paragraph{SOTA performance with MLLMs.}
As shown in Tables~\ref{tab:1}, \ref{tab:2}, and \ref{tab:3}, ADHint achieves consistent gains across multiple backbones, validating its robust generalization regarding model capabilities, scales, and families.
For Qwen2.5-VL, ADHint improves overall \textit{pass@1}/\textit{avg@8} over GRPO by 2.1\%/2.0\% at 7B and 2.3\%/2.1\% at 3B.
Compared with recent hint-based methods, ADHint achieves gains of 6.8\%/11.1\% (7B) and 9.7\%/10.5\% (3B).
For Qwen3-VL, ADHint outperforms the baselines by 5.1\%/0.8\% at 8B and 4.1\%/3.0\% at 4B.
Beyond the Qwen family, it achieves gains of 2.0\%/2.2\% over the baselines on MiMo-VL-7B-SFT-2508.
We observe that several baselines perform worse than vanilla GRPO, which is consistent with our motivation and the training dynamics observed. 
Methods using fixed, high hint ratios (e.g., GHPO and HintGRPO) can provide excessive guidance and drive training rewards above $0.9$, yet overfit to the hint distribution, limiting generalization and even degrading reasoning without hints.
Several methods (e.g., StepHint and LUFFY) ignore the imbalance in intra-group advantage estimation and the sharp gradients induced by hint tokens, producing unstable updates and training collapse.
These weaknesses are further exposed on the more challenging ViRL-Hint17k, whose reasoning chains average 3,100 tokens, versus typically a few hundred in prior baseline datasets (e.g., $\sim$230 tokens).
ADHint addresses these issues through novel mechanisms for hint-ratio scheduling, relative-advantage estimation, and hint-gradient modulation, providing a feasible solution for hint-based on-policy learning on challenging multimodal tasks.
More in-depth comparative analysis of ADHint and the baselines is provided in Appendix~\ref{Appendix:E}.
\paragraph{Performance with LLM.}
Table~\ref{tab:4} reports ADHint's performance on text-only math reasoning using the same data and experimental settings as GHPO~\cite{liu2025ghpo}.
Under this setting, ADHint outperforms the baselines by 2.4\% in average accuracy, showing that its effectiveness is not confined to the multimodal ViRL-Hint17k setting.
This result also indicates that ADHint can broaden model capability boundaries across modalities.
\paragraph{Generalization to different domains.}
Table~\ref{tab:5} reports results on the Medical VQA domain, which is knowledge-intensive, more OOD for the backbone models, and substantially different from standard math reasoning.
We observe that the baselines do not generalize well, whereas ADHint improves accuracy by 1.7\% over GRPO. These results suggest that ADHint can be readily transferred to more challenging real-world domains.
\subsection{Training Dynamics}
Figure~\ref{fig5} summarizes the training dynamics of key metrics.
Figure~\ref{fig5:sub_a} shows that ADHint maintains rewards within a moderate regime, yielding low-variance update signals. 
Figures~\ref{fig5:sub_b} and~\ref{fig5:sub_c} indicate that ADHint maintains exploration within a healthy entropy range with more stable gradient updates. 
Figures~\ref{fig5:sub_d} and~\ref{fig5:sub_e} show that ADHint steadily increases response length, while the proportion of overlength responses does not increase, which usually indicates repetition or garbled outputs. These trends suggest that the policy continuously acquires effective long-term thinking capabilities. 
Taken together, ADHint stably and adaptively learns reasoning ability from off-policy hints. 
\subsection{Ablation Study}
Table~\ref{tab:6} reports ablations of four critical components.
Specifically, ``w/o AH-SDP'' replaces adaptive hint-ratio schedule with an annealed strategy, while ``w/o AE-RDP'' pools hint- and naive-rollouts into a single group for advantage estimation. 
Removing any component degrades both \textit{pass@1} and \textit{avg@8}, underscoring the importance of our design.
The difficulty-aware design in the hint-ratio schedule and relative-advantage estimation ensures efficient learning that balances exploration and imitation. 
The gradient modulation mechanisms prevent biased and destructive updates, ensuring healthy policy learning. 
\subsection{Further Analysis}
\paragraph{A 20\% hint ratio is sufficient to guide reasoning.}
Table~\ref{tab:7} shows that performance peaks at $w_{\text{max}}=0.2$ and declines with either higher or lower hint ratios.
As shown in Appendix Figure~\ref{fig5:sub_a}, a 20\% hint ratio is sufficient to keep rewards within the range of 0.5--0.7, providing stable learning signals at a moderate difficulty. 
Higher hint ratios yield excessively high rewards for hint-rollouts, which introduces larger update bias. 
Above results suggest that the initial reasoning prefix is critical for steering the policy toward effective reasoning.
\paragraph{Hint-token gradients should be restricted to a reasonable range.}
Table~\ref{tab:8} evaluates the CGM threshold $\alpha$, which controls the gradient contributions of hint tokens, and shows that $\alpha=0.5$ performs best overall.
Including hint tokens with excessively high or low entropy in gradient updates shifts the policy markedly from its initial distribution, whereas an overly large threshold limits effective learning from off-policy hints and reduces the efficiency of knowledge expansion.
\section{Conclusion}
\label{sec:conclusion}
We propose ADHint, which explicitly incorporates difficulty into hint-based on-policy learning.
Enhanced by critical components that schedule hint ratios using the sample difficulty prior, estimate relative advantages using the rollout difficulty posterior, and modulate token-level gradients based on distribution consistency, ADHint surpasses existing methods across diverse settings, showcasing advanced reasoning and OOD generalization. 
\section*{Limitations}
Despite ADHint's strong performance across diverse modalities, model scales, model families, and domains, several directions remain for future investigation.
First, although we include an additional experiment on a 32B backbone in Appendix~\ref{Appendix:B.32B}, our empirical evaluation primarily focuses on models up to the 8B scale.
Systematic experiments on models with 32B or more parameters would provide a more comprehensive understanding of ADHint's scaling behavior.
Second, our experiments mainly consider reasoning tasks with verifiable answers, which enable reliable outcome-based evaluation and reinforcement learning.
Exploring the generalization of ADHint to open-ended tasks without readily verifiable answers represents another promising direction for future work.
\section*{Ethical Considerations}
This work studies reinforcement learning methods for improving the reasoning abilities of LLMs and MLLMs. 
Our training data are derived from publicly available datasets, and we do not collect new data or intentionally include private, sensitive, or personally identifiable information. 
The broader risks of ADHint are similar to those associated with general-purpose reasoning models, including misuse, unreliable outputs, and insufficiently supervised deployment. 
We emphasize the importance of careful evaluation and appropriate safeguards to ensure fair, safe, and responsible use of LLM and MLLM systems.
\section*{Acknowledgement}
This work was supported by Qwen Business Unit through Alibaba Research Intern Program.

\FloatBarrier
\bibliography{custom}

@article{wang2025vl,
  title={Vl-rethinker: Incentivizing self-reflection of vision-language models with reinforcement learning},
  author={Wang, Haozhe and Qu, Chao and Huang, Zuming and Chu, Wei and Lin, Fangzhen and Chen, Wenhu},
  journal={arXiv preprint arXiv:2504.08837},
  year={2025}
}

@misc{bai2025qwen3vltechnicalreport,
      title={Qwen3-VL Technical Report}, 
      author={Shuai Bai and Yuxuan Cai and Ruizhe Chen and Keqin Chen and Xionghui Chen and Zesen Cheng and Lianghao Deng and Wei Ding and Chang Gao and Chunjiang Ge and Wenbin Ge and Zhifang Guo and Qidong Huang and Jie Huang and Fei Huang and Binyuan Hui and Shutong Jiang and Zhaohai Li and Mingsheng Li and Mei Li and Kaixin Li and Zicheng Lin and Junyang Lin and Xuejing Liu and Jiawei Liu and Chenglong Liu and Yang Liu and Dayiheng Liu and Shixuan Liu and Dunjie Lu and Ruilin Luo and Chenxu Lv and Rui Men and Lingchen Meng and Xuancheng Ren and Xingzhang Ren and Sibo Song and Yuchong Sun and Jun Tang and Jianhong Tu and Jianqiang Wan and Peng Wang and Pengfei Wang and Qiuyue Wang and Yuxuan Wang and Tianbao Xie and Yiheng Xu and Haiyang Xu and Jin Xu and Zhibo Yang and Mingkun Yang and Jianxin Yang and An Yang and Bowen Yu and Fei Zhang and Hang Zhang and Xi Zhang and Bo Zheng and Humen Zhong and Jingren Zhou and Fan Zhou and Jing Zhou and Yuanzhi Zhu and Ke Zhu},
      year={2025},
      eprint={2511.21631},
      archivePrefix={arXiv},
      primaryClass={cs.CV},
      url={https://arxiv.org/abs/2511.21631}, 
}

@article{bai2025qwen2,
  title={Qwen2.5-vl technical report},
  author={Bai, Shuai and Chen, Keqin and Liu, Xuejing and Wang, Jialin and Ge, Wenbin and Song, Sibo and Dang, Kai and Wang, Peng and Wang, Shijie and Tang, Jun and others},
  journal={arXiv preprint arXiv:2502.13923},
  year={2025}
}

@article{lu2023mathvista,
  title={Mathvista: Evaluating math reasoning in visual contexts with gpt-4v, bard, and other large multimodal models},
  author={Lu, Pan and Bansal, Hritik and Xia, Tony and Liu, Jiacheng and Li, Chunyuan and Hajishirzi, Hannaneh and Cheng, Hao and Chang, Kai-Wei and Galley, Michel and Gao, Jianfeng},
  journal={CoRR},
  year={2023}
}

@inproceedings{zhang2024mathverse,
  title={Mathverse: Does your multi-modal llm truly see the diagrams in visual math problems?},
  author={Zhang, Renrui and Jiang, Dongzhi and Zhang, Yichi and Lin, Haokun and Guo, Ziyu and Qiu, Pengshuo and Zhou, Aojun and Lu, Pan and Chang, Kai-Wei and Qiao, Yu and others},
  booktitle={European Conference on Computer Vision},
  pages={169--186},
  year={2024},
  organization={Springer}
}

@article{qiao2024we,
  title={We-math: Does your large multimodal model achieve human-like mathematical reasoning?},
  author={Qiao, Runqi and Tan, Qiuna and Dong, Guanting and Wu, Minhui and Sun, Chong and Song, Xiaoshuai and GongQue, Zhuoma and Lei, Shanglin and Wei, Zhe and Zhang, Miaoxuan and others},
  journal={arXiv preprint arXiv:2407.01284},
  year={2024}
}

@inproceedings{yue2024mmmu,
  title={Mmmu: A massive multi-discipline multimodal understanding and reasoning benchmark for expert agi},
  author={Yue, Xiang and Ni, Yuansheng and Zhang, Kai and Zheng, Tianyu and Liu, Ruoqi and Zhang, Ge and Stevens, Samuel and Jiang, Dongfu and Ren, Weiming and Sun, Yuxuan and others},
  booktitle={Proceedings of the IEEE/CVF Conference on Computer Vision and Pattern Recognition},
  pages={9556--9567},
  year={2024}
}

@article{yue2024mmmupro,
  title={Mmmu-pro: A more robust multi-discipline multimodal understanding benchmark},
  author={Yue, Xiang and Zheng, Tianyu and Ni, Yuansheng and Wang, Yubo and Zhang, Kai and Tong, Shengbang and Sun, Yuxuan and Yu, Botao and Zhang, Ge and Sun, Huan and others},
  journal={arXiv preprint arXiv:2409.02813},
  year={2024}
}

@article{xiao2024logicvista,
  title={Logicvista: Multimodal llm logical reasoning benchmark in visual contexts},
  author={Xiao, Yijia and Sun, Edward and Liu, Tianyu and Wang, Wei},
  journal={arXiv preprint arXiv:2407.04973},
  year={2024}
}

@article{shao2024deepseekmath,
  title={Deepseekmath: Pushing the limits of mathematical reasoning in open language models},
  author={Shao, Zhihong and Wang, Peiyi and Zhu, Qihao and Xu, Runxin and Song, Junxiao and Bi, Xiao and Zhang, Haowei and Zhang, Mingchuan and Li, YK and others},
  journal={arXiv preprint arXiv:2402.03300},
  year={2024}
}

@inproceedings{yan2025learning,
title={Learning to Reason under Off-Policy Guidance},
author={Jianhao Yan and Yafu Li and Zican Hu and Zhi Wang and Ganqu Cui and Xiaoye Qu and Yu Cheng and Yue Zhang},
booktitle={The Thirty-ninth Annual Conference on Neural Information Processing Systems},
year={2025},
url={https://openreview.net/forum?id=vO8LLoNWWk}
}

@article{ma2025learning,
  title={Learning What Reinforcement Learning Can't: Interleaved Online Fine-Tuning for Hardest Questions},
  author={Ma, Lu and Liang, Hao and Qiang, Meiyi and Tang, Lexiang and Ma, Xiaochen and Wong, Zhen Hao and Niu, Junbo and Shen, Chengyu and He, Runming and Li, Yanhao and others},
  journal={arXiv preprint arXiv:2506.07527},
  year={2025}
}

@article{dong2025rl,
  title={Rl-plus: Countering capability boundary collapse of llms in reinforcement learning with hybrid-policy optimization},
  author={Dong, Yihong and Jiang, Xue and Tao, Yongding and Liu, Huanyu and Zhang, Kechi and Mou, Lili and Cao, Rongyu and Ma, Yingwei and Chen, Jue and Li, Binhua and others},
  journal={arXiv preprint arXiv:2508.00222},
  year={2025}
}

@InProceedings{Huang_2025_ICCV,
    author    = {Huang, Qihan and Dai, Weilong and Liu, Jinlong and He, Wanggui and Jiang, Hao and Song, Mingli and Chen, Jingyuan and Yao, Chang and Song, Jie},
    title     = {Boosting MLLM Reasoning with Text-Debiased Hint-GRPO},
    booktitle = {Proceedings of the IEEE/CVF International Conference on Computer Vision (ICCV)},
    month     = {October},
    year      = {2025},
    pages     = {4848-4857}
}

@inproceedings{liu2025uft,
title={{UFT}: Unifying Supervised and Reinforcement Fine-Tuning},
author={Mingyang Liu and Gabriele Farina and Asuman E. Ozdaglar},
booktitle={The Thirty-ninth Annual Conference on Neural Information Processing Systems},
year={2025},
url={https://openreview.net/forum?id=usOkGv1S7M}
}

@article{liu2025ghpo,
  title={Ghpo: Adaptive guidance for stable and efficient llm reinforcement learning},
  author={Liu, Ziru and Gong, Cheng and Fu, Xinyu and Liu, Yaofang and Chen, Ran and Hu, Shoubo and Zhang, Suiyun and Liu, Rui and Zhang, Qingfu and Tu, Dandan},
  journal={arXiv preprint arXiv:2507.10628},
  year={2025}
}

@article{guo2025deepseek,
  title={Deepseek-r1 incentivizes reasoning in llms through reinforcement learning},
  author={Guo, Daya and Yang, Dejian and Zhang, Haowei and Song, Junxiao and Wang, Peiyi and Zhu, Qihao and Xu, Runxin and Zhang, Ruoyu and Ma, Shirong and Bi, Xiao and others},
  journal={Nature},
  volume={645},
  number={8081},
  pages={633--638},
  year={2025},
  publisher={Nature Publishing Group UK London}
}

@article{li2024numinamath,
  title={Numinamath: The largest public dataset in ai4maths with 860k pairs of competition math problems and solutions},
  author={Li, Jia and Beeching, Edward and Tunstall, Lewis and Lipkin, Ben and Soletskyi, Roman and Huang, Shengyi and Rasul, Kashif and Yu, Longhui and Jiang, Albert Q and Shen, Ziju and others},
  journal={Hugging Face repository},
  volume={13},
  number={9},
  pages={9},
  year={2024}
}

@article{lewkowycz2022solving,
  title={Solving quantitative reasoning problems with language models},
  author={Lewkowycz, Aitor and Andreassen, Anders and Dohan, David and Dyer, Ethan and Michalewski, Henryk and Ramasesh, Vinay and Slone, Ambrose and Anil, Cem and Schlag, Imanol and Gutman-Solo, Theo and others},
  journal={Advances in neural information processing systems},
  volume={35},
  pages={3843--3857},
  year={2022}
}

@article{he2024olympiadbench,
  title={Olympiadbench: A challenging benchmark for promoting agi with olympiad-level bilingual multimodal scientific problems},
  author={He, Chaoqun and Luo, Renjie and Bai, Yuzhuo and Hu, Shengding and Thai, Zhen Leng and Shen, Junhao and Hu, Jinyi and Han, Xu and Huang, Yujie and Zhang, Yuxiang and others},
  journal={arXiv preprint arXiv:2402.14008},
  year={2024}
}

@inproceedings{hendrycks2measuring,
  title={Measuring Mathematical Problem Solving With the MATH Dataset},
  author={Hendrycks, Dan and Burns, Collin and Kadavath, Saurav and Arora, Akul and Basart, Steven and Tang, Eric and Song, Dawn and Steinhardt, Jacob},
  booktitle={Thirty-fifth Conference on Neural Information Processing Systems Datasets and Benchmarks Track (Round 2)},
  year={2021}
}

@article{yang2024qwen2,
  title={Qwen2. 5-math technical report: Toward mathematical expert model via self-improvement},
  author={Yang, An and Zhang, Beichen and Hui, Binyuan and Gao, Bofei and Yu, Bowen and Li, Chengpeng and Liu, Dayiheng and Tu, Jianhong and Zhou, Jingren and Lin, Junyang and others},
  journal={arXiv preprint arXiv:2409.12122},
  year={2024}
}

@inproceedings{chusft,
  title={SFT Memorizes, RL Generalizes: A Comparative Study of Foundation Model Post-training},
  author={Chu, Tianzhe and Zhai, Yuexiang and Yang, Jihan and Tong, Shengbang and Xie, Saining and Schuurmans, Dale and Le, Quoc V and Levine, Sergey and Ma, Yi},
  booktitle={Forty-second International Conference on Machine Learning},
  year={2025}
}

@article{li2025system,
  title={From system 1 to system 2: A survey of reasoning large language models},
  author={Li, Zhong-Zhi and Zhang, Duzhen and Zhang, Ming-Liang and Zhang, Jiaxin and Liu, Zengyan and Yao, Yuxuan and Xu, Haotian and Zheng, Junhao and Wang, Pei-Jie and Chen, Xiuyi and others},
  journal={arXiv preprint arXiv:2502.17419},
  year={2025}
}

@article{zeng2025glm,
  title={Glm-4.5: Agentic, reasoning, and coding (arc) foundation models},
  author={Zeng, Aohan and Lv, Xin and Zheng, Qinkai and Hou, Zhenyu and Chen, Bin and Xie, Chengxing and Wang, Cunxiang and Yin, Da and Zeng, Hao and Zhang, Jiajie and others},
  journal={arXiv preprint arXiv:2508.06471},
  year={2025}
}

@article{yang2025qwen3,
  title={Qwen3 technical report},
  author={Yang, An and Li, Anfeng and Yang, Baosong and Zhang, Beichen and Hui, Binyuan and Zheng, Bo and Yu, Bowen and Gao, Chang and Huang, Chengen and Lv, Chenxu and others},
  journal={arXiv preprint arXiv:2505.09388},
  year={2025}
}

@article{liu2025visual,
  title={Visual-rft: Visual reinforcement fine-tuning},
  author={Liu, Ziyu and Sun, Zeyi and Zang, Yuhang and Dong, Xiaoyi and Cao, Yuhang and Duan, Haodong and Lin, Dahua and Wang, Jiaqi},
  journal={arXiv preprint arXiv:2503.01785},
  year={2025}
}

@article{huang2025vision,
  title={Vision-r1: Incentivizing reasoning capability in multimodal large language models},
  author={Huang, Wenxuan and Jia, Bohan and Zhai, Zijie and Cao, Shaosheng and Ye, Zheyu and Zhao, Fei and Xu, Zhe and Hu, Yao and Lin, Shaohui},
  journal={arXiv preprint arXiv:2503.06749},
  year={2025}
}

@article{fu2025srft,
  title={SRFT: A Single-Stage Method with Supervised and Reinforcement Fine-Tuning for Reasoning},
  author={Fu, Yuqian and Chen, Tinghong and Chai, Jiajun and Wang, Xihuai and Tu, Songjun and Yin, Guojun and Lin, Wei and Zhang, Qichao and Zhu, Yuanheng and Zhao, Dongbin},
  journal={arXiv preprint arXiv:2506.19767},
  year={2025}
}

@article{zhang2025bread,
  title={BREAD: Branched Rollouts from Expert Anchors Bridge SFT \& RL for Reasoning},
  author={Zhang, Xuechen and Huang, Zijian and Li, Yingcong and Ni, Chenshun and Chen, Jiasi and Oymak, Samet},
  journal={arXiv preprint arXiv:2506.17211},
  year={2025}
}

@article{chen2025data,
  title={From data-centric to sample-centric: Enhancing llm reasoning via progressive optimization},
  author={Chen, Xinjie and Liao, Minpeng and Chen, Guoxin and Li, Chengxi and Fu, Biao and Fan, Kai and Liu, Xinggao},
  journal={arXiv preprint arXiv:2507.06573},
  year={2025}
}

@article{zhang2025stephint,
  title={StepHint: Multi-level Stepwise Hints Enhance Reinforcement Learning to Reason},
  author={Zhang, Kaiyi and Lv, Ang and Li, Jinpeng and Wang, Yongbo and Wang, Feng and Hu, Haoyuan and Yan, Rui},
  journal={arXiv preprint arXiv:2507.02841},
  year={2025}
}

@article{huang2025blending,
  title={Blending Supervised and Reinforcement Fine-Tuning with Prefix Sampling},
  author={Huang, Zeyu and Cheng, Tianhao and Qiu, Zihan and Wang, Zili and Xu, Yinghui and Ponti, Edoardo M and Titov, Ivan},
  journal={arXiv preprint arXiv:2507.01679},
  year={2025}
}

@article{schulman2017proximal,
  title={Proximal policy optimization algorithms},
  author={Schulman, John and Wolski, Filip and Dhariwal, Prafulla and Radford, Alec and Klimov, Oleg},
  journal={arXiv preprint arXiv:1707.06347},
  year={2017}
}

@article{tan2025towards,
  title={Towards Flash Thinking via Decoupled Advantage Policy Optimization},
  author={Tan, Zezhong and Gao, Hang and Ma, Xinhong and Zhang, Feng and Dong, Ziqiang},
  journal={arXiv preprint arXiv:2510.15374},
  year={2025}
}

@article{team2025kimi,
  title={Kimi k2: Open agentic intelligence},
  author={Team, Kimi and Bai, Yifan and Bao, Yiping and Chen, Guanduo and Chen, Jiahao and Chen, Ningxin and Chen, Ruijue and Chen, Yanru and Chen, Yuankun and Chen, Yutian and others},
  journal={arXiv preprint arXiv:2507.20534},
  year={2025}
}

@article{zheng2025deepeyes,
  title={DeepEyes: Incentivizing ``Thinking with Images'' via Reinforcement Learning},
  author={Zheng, Ziwei and Yang, Michael and Hong, Jack and Zhao, Chenxiao and Xu, Guohai and Yang, Le and Shen, Chao and Yu, Xing},
  journal={arXiv preprint arXiv:2505.14362},
  year={2025}
}

@article{su2025pixel,
  title={Pixel reasoner: Incentivizing pixel-space reasoning with curiosity-driven reinforcement learning},
  author={Su, Alex and Wang, Haozhe and Ren, Weiming and Lin, Fangzhen and Chen, Wenhu},
  journal={arXiv preprint arXiv:2505.15966},
  year={2025}
}

@article{liu2025noisyrollout,
  title={Noisyrollout: Reinforcing visual reasoning with data augmentation},
  author={Liu, Xiangyan and Ni, Jinjie and Wu, Zijian and Du, Chao and Dou, Longxu and Wang, Haonan and Pang, Tianyu and Shieh, Michael Qizhe},
  journal={arXiv preprint arXiv:2504.13055},
  year={2025}
}

@article{wang2025perception,
  title={Perception-aware policy optimization for multimodal reasoning},
  author={Wang, Zhenhailong and Guo, Xuehang and Stoica, Sofia and Xu, Haiyang and Wang, Hongru and Ha, Hyeonjeong and Chen, Xiusi and Chen, Yangyi and Yan, Ming and Huang, Fei and others},
  journal={arXiv preprint arXiv:2507.06448},
  year={2025}
}

@article{liu2025visionreasoner,
  title={VisionReasoner: Unified Visual Perception and Reasoning via Reinforcement Learning},
  author={Liu, Yuqi and Qu, Tianyuan and Zhong, Zhisheng and Peng, Bohao and Liu, Shu and Yu, Bei and Jia, Jiaya},
  journal={arXiv preprint arXiv:2505.12081},
  year={2025}
}

@misc{liang2025modomodomultidomaindatamixtures,
      title={MoDoMoDo: Multi-Domain Data Mixtures for Multimodal LLM Reinforcement Learning}, 
      author={Yiqing Liang and Jielin Qiu and Wenhao Ding and Zuxin Liu and James Tompkin and Mengdi Xu and Mengzhou Xia and Zhengzhong Tu and Laixi Shi and Jiacheng Zhu},
      year={2025},
      eprint={2505.24871},
      archivePrefix={arXiv},
      primaryClass={cs.CV},
      url={https://arxiv.org/abs/2505.24871}, 
}

@inproceedings{bengio2009curriculum,
  title={Curriculum learning},
  author={Bengio, Yoshua and Louradour, J{\'e}r{\^o}me and Collobert, Ronan and Weston, Jason},
  booktitle={Proceedings of the 26th annual international conference on machine learning},
  pages={41--48},
  year={2009}
}

@article{parashar2025curriculum,
  title={Curriculum Reinforcement Learning from Easy to Hard Tasks Improves LLM Reasoning},
  author={Parashar, Shubham and Gui, Shurui and Li, Xiner and Ling, Hongyi and Vemuri, Sushil and Olson, Blake and Li, Eric and Zhang, Yu and Caverlee, James and Kalathil, Dileep and others},
  journal={arXiv preprint arXiv:2506.06632},
  year={2025}
}

@misc{xi2025bapostabilizingoffpolicyreinforcement,
      title={BAPO: Stabilizing Off-Policy Reinforcement Learning for LLMs via Balanced Policy Optimization with Adaptive Clipping}, 
      author={Zhiheng Xi and Xin Guo and Yang Nan and Enyu Zhou and Junrui Shen and Wenxiang Chen and Jiaqi Liu and Jixuan Huang and Zhihao Zhang and Honglin Guo and Xun Deng and Zhikai Lei and Miao Zheng and Guoteng Wang and Shuo Zhang and Peng Sun and Rui Zheng and Hang Yan and Tao Gui and Qi Zhang and Xuanjing Huang},
      year={2025},
      eprint={2510.18927},
      archivePrefix={arXiv},
      primaryClass={cs.LG},
      url={https://arxiv.org/abs/2510.18927}, 
}

@misc{wu2025templaterlstructuredtemplateguidedreinforcement,
      title={TemplateRL: Structured Template-Guided Reinforcement Learning for LLM Reasoning}, 
      author={Jinyang Wu and Chonghua Liao and Mingkuan Feng and Shuai Zhang and Zhengqi Wen and Haoran Luo and Ling Yang and Huazhe Xu and Jianhua Tao},
      year={2025},
      eprint={2505.15692},
      archivePrefix={arXiv},
      primaryClass={cs.CL},
      url={https://arxiv.org/abs/2505.15692}, 
}

@misc{zheng2025easyr1,
  title={EasyR1: An Efficient, Scalable, Multi-Modality RL Training Framework},
  author= {Yaowei Zheng and Junting Lu and Shenzhi Wang and Zhangchi Feng and Dongdong Kuang and Yuwen Xiong},
  howpublished = {https://github.com/hiyouga/EasyR1},
  year         = {2025}
}

@misc{coreteam2025mimovltechnicalreport,
      title={MiMo-VL Technical Report}, 
      author={LLM-Core-Team Xiaomi},
      year={2025},
      eprint={2506.03569},
      archivePrefix={arXiv},
      primaryClass={cs.CL},
      url={https://arxiv.org/abs/2506.03569}, 
}

@article{zhang2023pmc,
  title={Pmc-vqa: Visual instruction tuning for medical visual question answering},
  author={Zhang, Xiaoman and Wu, Chaoyi and Zhao, Ziheng and Lin, Weixiong and Zhang, Ya and Wang, Yanfeng and Xie, Weidi},
  journal={arXiv preprint arXiv:2305.10415},
  year={2023}
}

@article{patel2024aime,
  title={Aime: Ai system optimization via multiple llm evaluators},
  author={Patel, Bhrij and Chakraborty, Souradip and Suttle, Wesley A and Wang, Mengdi and Bedi, Amrit Singh and Manocha, Dinesh},
  journal={arXiv preprint arXiv:2410.03131},
  year={2024}
}

@article{li2025staying,
  title={Staying in the Sweet Spot: Responsive Reasoning Evolution via Capability-Adaptive Hint Scaffolding},
  author={Li, Ziheng and Sun, Zexu and Zhao, Jinman and Min, Erxue and Zeng, Yongcheng and Wu, Hui and Cai, Hengyi and Wang, Shuaiqiang and Yin, Dawei and Chen, Xu and others},
  journal={arXiv preprint arXiv:2509.06923},
  year={2025}
}

@article{li2025adacurl,
  title={Adacurl: Adaptive curriculum reinforcement learning with invalid sample mitigation and historical revisiting},
  author={Li, Renda and Huang, Hailang and Wei, Fei and Xiong, Feng and Wang, Yong and Chu, Xiangxiang},
  journal={arXiv preprint arXiv:2511.09478},
  year={2025}
}

@article{zhu2025toward,
  title={Toward effective reinforcement learning fine-tuning for medical vqa in vision-language models},
  author={Zhu, Wenhui and Dong, Xuanzhao and Li, Xin and Qiu, Peijie and Chen, Xiwen and Razi, Abolfazl and Sotiras, Aris and Su, Yi and Wang, Yalin},
  journal={arXiv preprint arXiv:2505.13973},
  year={2025}
}

\appendix
\section*{Overview}
\begin{figure*}[t]
	\centering
	\captionsetup{skip=4pt}
	\captionsetup[subfigure]{skip=2pt}
	
	\begin{subfigure}[t]{0.32\textwidth}
		\centering
		\includegraphics[width=\linewidth]{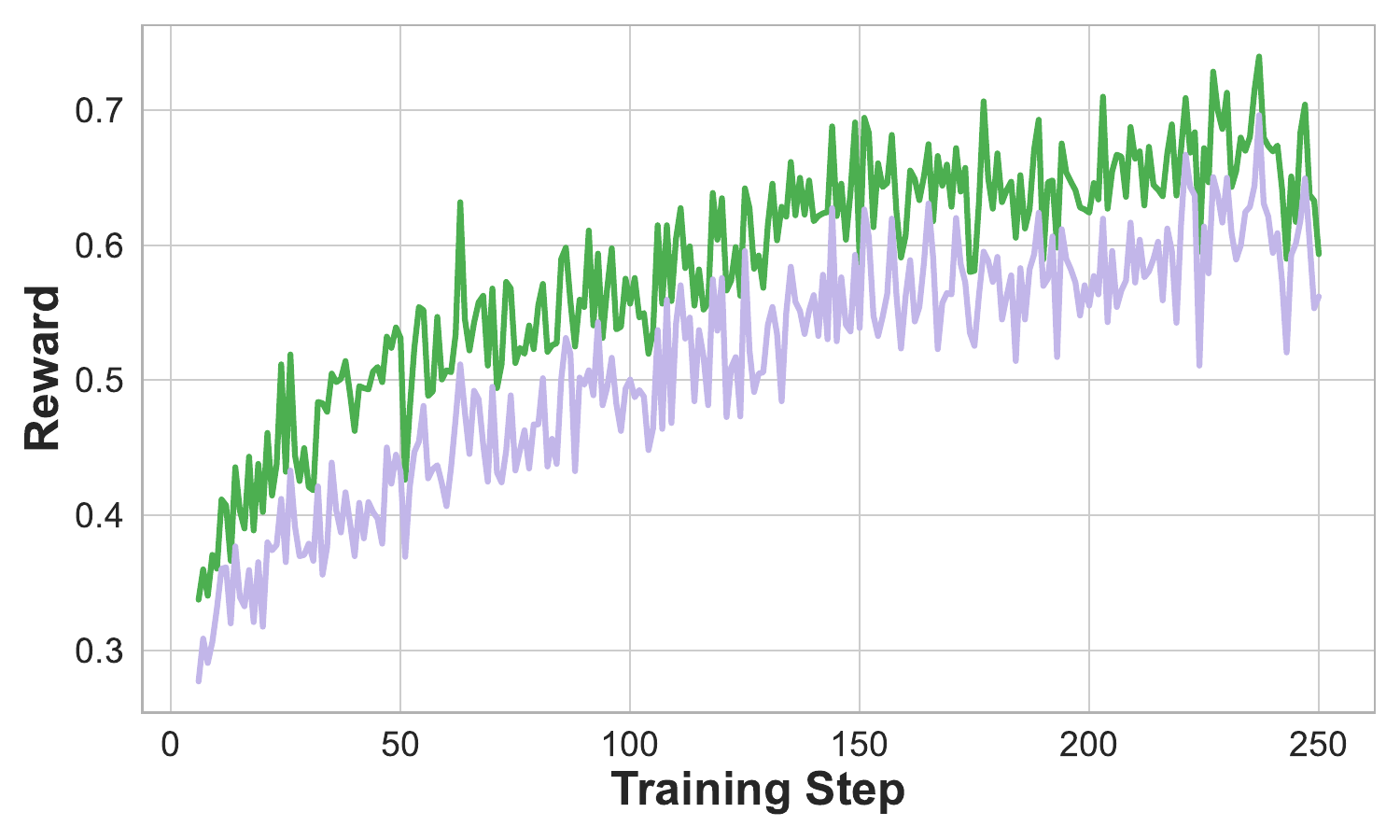}
		\caption{Training reward}
		\label{fig5:sub_a}
	\end{subfigure}
	\hfill
	\begin{subfigure}[t]{0.32\textwidth}
		\centering
		\includegraphics[width=\linewidth]{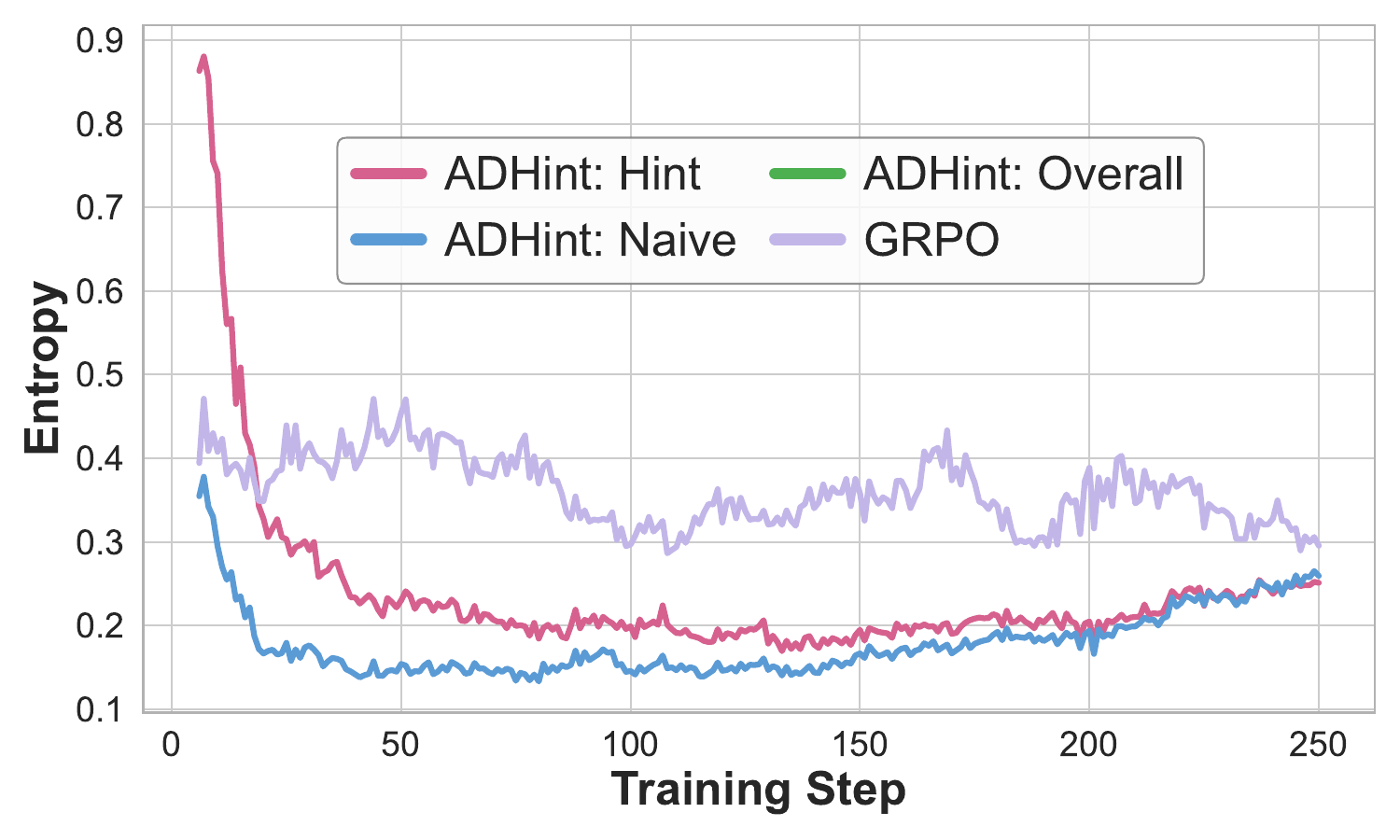}
		\caption{Policy entropy}
		\label{fig5:sub_b}
	\end{subfigure}
	\hfill
	\begin{subfigure}[t]{0.32\textwidth}
		\centering
		\includegraphics[width=\linewidth]{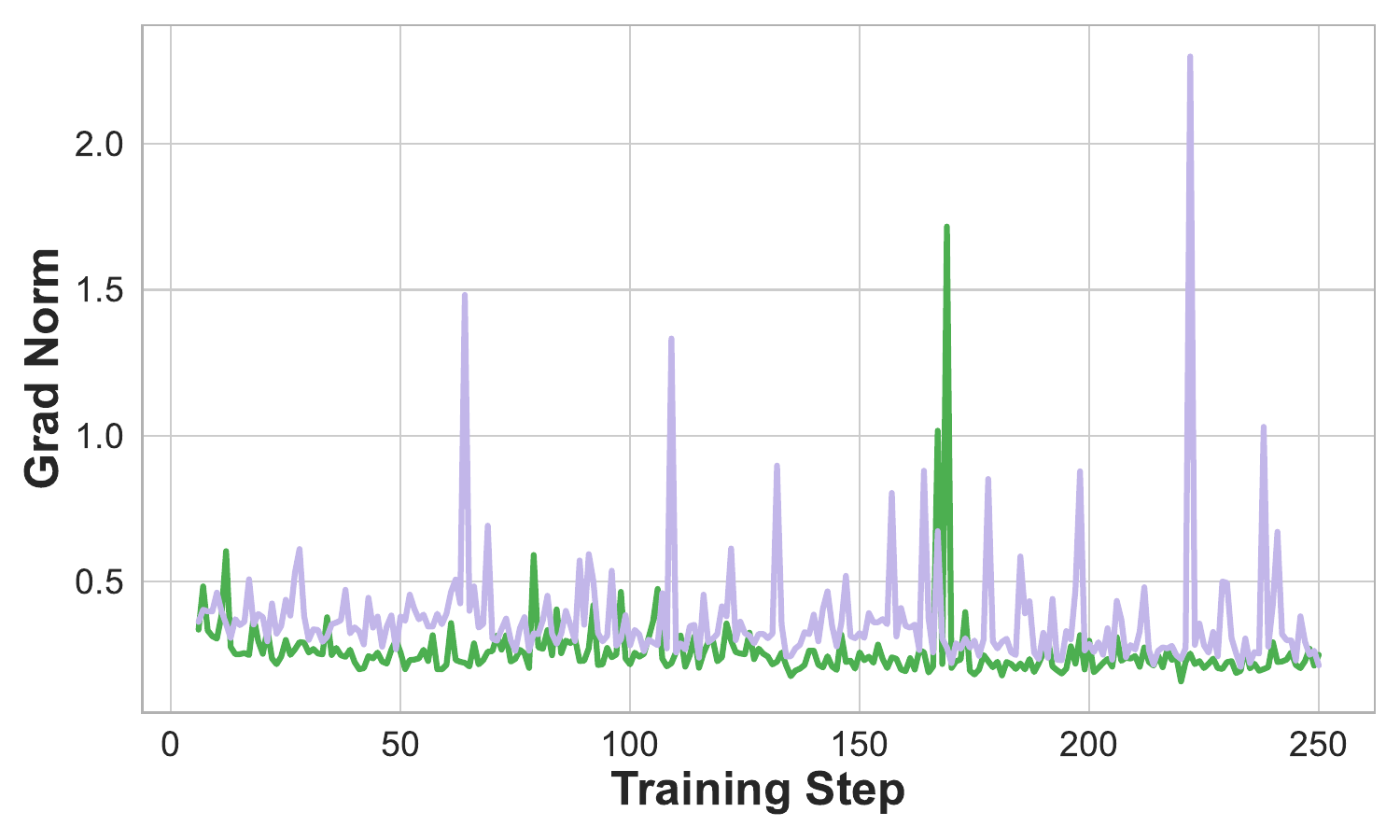}
		\caption{Gradient norm}
		\label{fig5:sub_c}
	\end{subfigure}
	\par\vspace{0.2em}
	
	\begin{subfigure}[t]{0.32\textwidth}
		\centering
		\includegraphics[width=\linewidth]{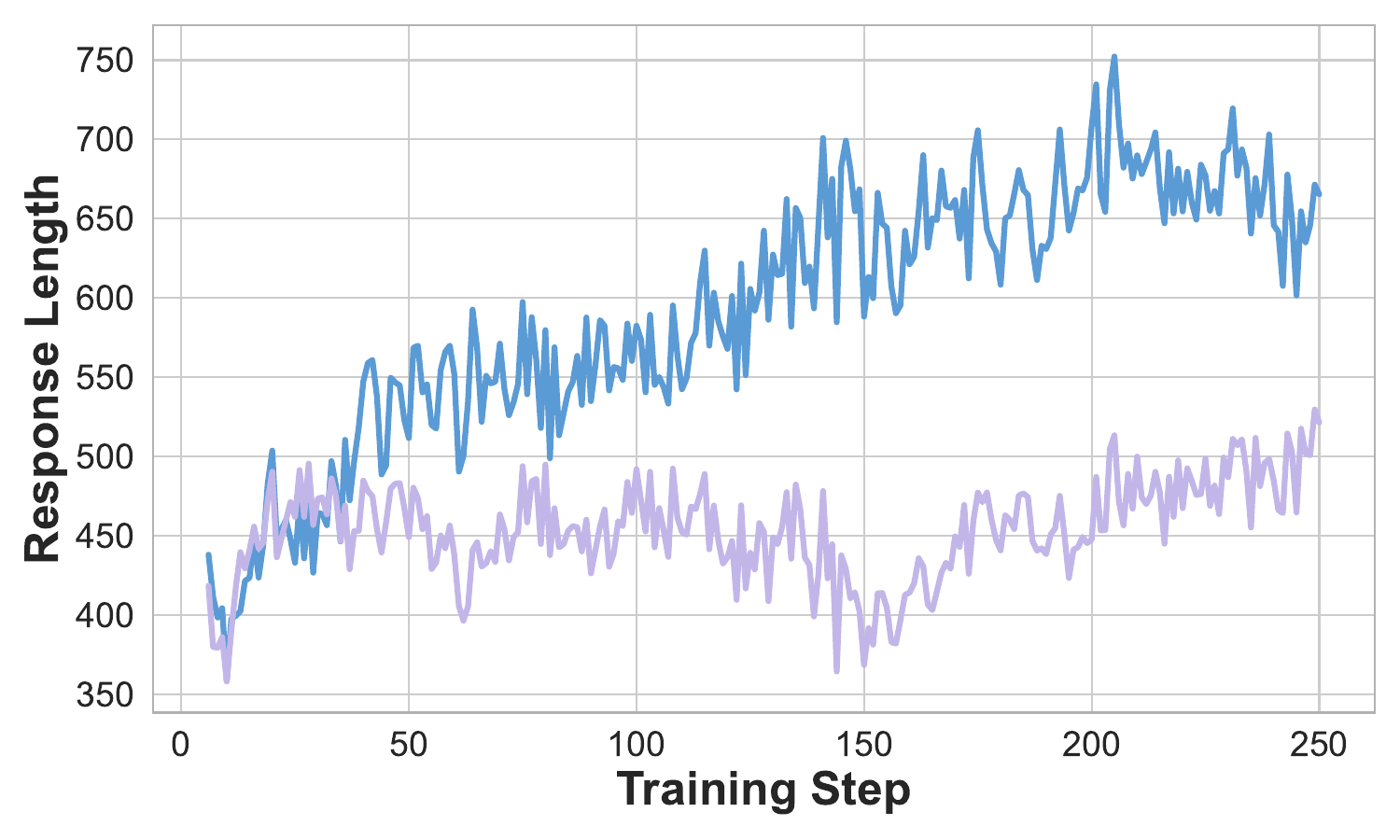}
		\caption{Response length}
		\label{fig5:sub_d}
	\end{subfigure}
	\hfill
	\begin{subfigure}[t]{0.32\textwidth}
		\centering
		\includegraphics[width=\linewidth]{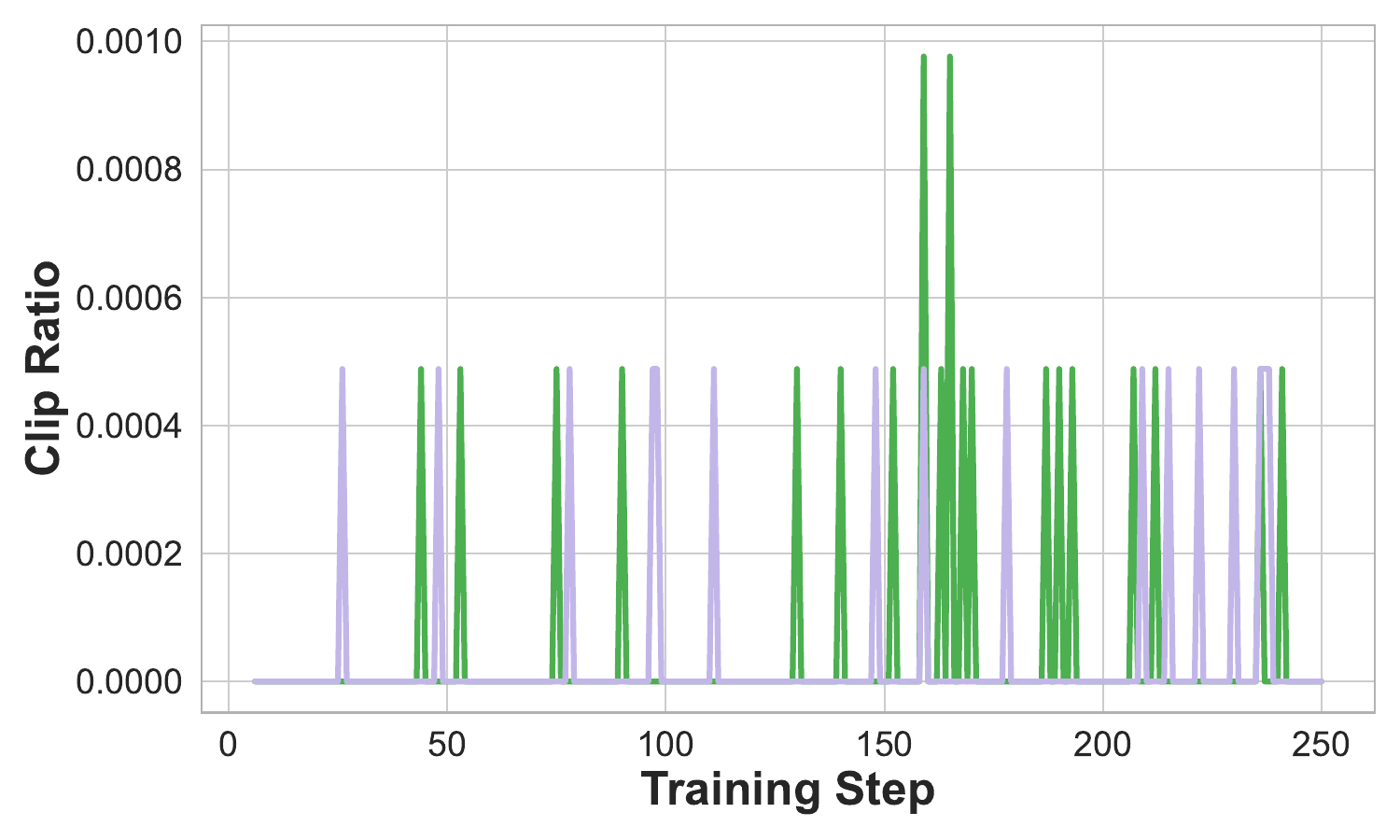}
		\caption{Clipping ratio of responses}
		\label{fig5:sub_e}
	\end{subfigure}
	\hfill
	\begin{subfigure}[t]{0.32\textwidth}
		\centering
		\includegraphics[width=\linewidth]{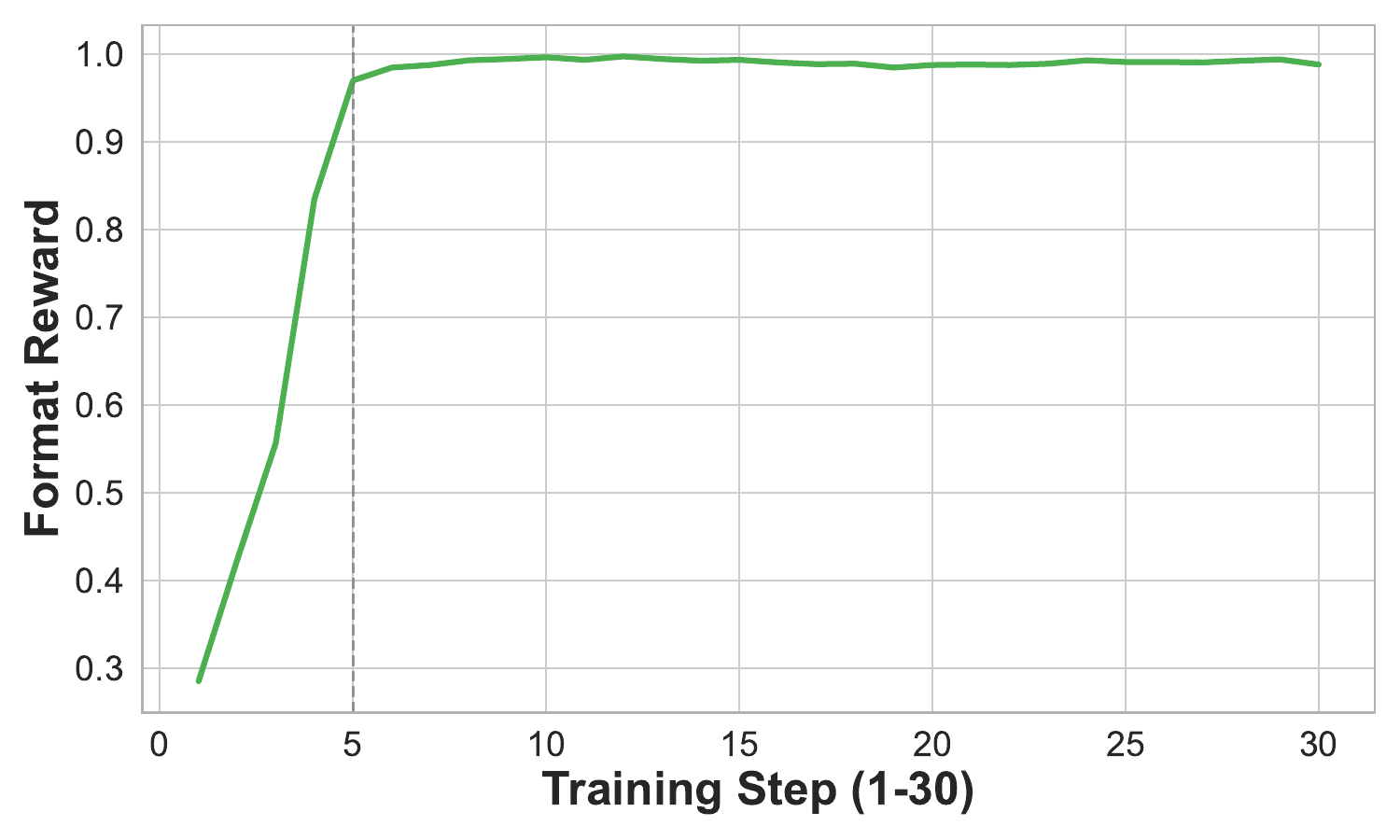}
		\caption{Format reward}
		\label{fig5:sub_f}
	\end{subfigure}
	\caption{
		\textbf{Training dynamics of key metrics for ADHint and GRPO.}
		``ADHint: Hint/Naive'' reports metrics computed on hint- or naive-rollouts only, while ``ADHint: Overall'' aggregates both rollout types. 
		ADHint maintains stable learning and yields concurrent improvements in reasoning generalization and knowledge expansion.
	}
	\label{fig5}
	\vspace{-0.4em}
\end{figure*}

Our appendix is organized as follows:
\begin{itemize}[leftmargin=*]
    \item \textbf{Section~\ref{Appendix:A}} presents the workflow of ADHint.
    \item \textbf{Section~\ref{Appendix:B}} presents the complete results of the generalization experiments reported in the main text on MiMo-VL-7B-SFT-2508 (Table~\ref{tab:3}), Qwen2.5-Math-7B (Table~\ref{tab:4}), and Medical VQA (Table~\ref{tab:5}), together with the training dynamics (Figure~\ref{fig5}) and analyses of the maximum hint ratio $w_{\text{max}}$ and CGM threshold $\alpha$ (Tables~\ref{tab:7} and~\ref{tab:8}). It further provides results at the 32B scale (Table~\ref{tab:32b}), comparisons with an additional supervised baseline (Table~\ref{tab:sft_variants}), and an analysis of computational overhead.
    \item \textbf{Section~\ref{Appendix:C}} provides an extended discussion of related work.
    \item \textbf{Section~\ref{Appendix:D}} offers additional details on training data construction, experimental settings, and the reimplementation of baselines.
    \item \textbf{Section~\ref{Appendix:E}} conducts an in-depth comparison between ADHint and the baselines, analyzing the sources of their performance differences.
    \item \textbf{Section~\ref{Appendix:F}} provides several case studies to offer a more intuitive view of ADHint's behavior.
\end{itemize}
\section{Training Algorithm of ADHint}
\label{Appendix:A}
In Algorithm~\ref{algo}, we provide the detailed workflow of ADHint as a complement to the main text. 
\begin{algorithm}[htbp]
    \caption{\textbf{ADHint}: \textbf{AD}aptive \textbf{Hint}s with Difficulty Priors for Reinforcement Learning}
    \label{algo}
    \footnotesize
    \begin{algorithmic}
    \REQUIRE Initial policy model $\pi_{\theta}$, training set $\mathcal{D}$, hyperparameters $w_{\text{max}}, \alpha$
    \ENSURE Optimized policy model $\pi_{\theta}^*$
    \FOR{$\text{step} = 1,\ldots,S$}
        \STATE Sample a mini-batch $\mathcal{D}_b \subset \mathcal{D}$
        \FOR{each $q \in \mathcal{D}_b$}
            \STATE Generate naive-rollouts $o_{1:n} \sim \pi_{\theta}(\cdot \mid q)$
            \STATE Compute $\text{Diff}_{\text{N}}$ by Eq.~\ref{eq:diff_naive}
            \STATE Compute hint ratio $w$ by Eq.~\ref{eq:hint_ratio}
            \STATE Generate hint-rollouts $o_{n+1:n+m} \sim \pi_{\theta}(\cdot \mid q, o_{i,<h})$
            \STATE Compute $\text{Diff}_{\text{H}}$ by Eq.~\ref{eq:diff_hint}
            \STATE Compute relative advantages $\tilde{A}_i$ by Eq.~\ref{eq:adv_rdp}
            \FOR{each hint-rollout $o_i \in \{o_{n+1},\ldots,o_{n+m}\}$}
                \STATE Compute token entropy by Eq.~\ref{eq:token_entropy}
                \STATE Compute continuation entropy by Eq.~\ref{eq:seq_entropy} 
                \STATE Compute token-wise gradient modulation factors $k_{i,t}$ by Eq.~\ref{eq:factor}
            \ENDFOR
        \ENDFOR
        \STATE Update $\theta$ by ascending $\mathcal{J}_{\text{ADHint}}^*(\pi_{\theta})$ in Eq.~\ref{eq:final_grad}
    \ENDFOR
    \RETURN $\pi_{\theta}^*$
    \end{algorithmic}
\end{algorithm}
\section{Additional Experimental Results}
\label{Appendix:B}
\subsection{Results on Larger Models}
\label{Appendix:B.32B}
Table~\ref{tab:32b} reports results on Qwen2.5-VL-32B.
ADHint achieves the best overall \textit{pass@1}/\textit{avg@8} of 68.1\%/68.2\%, surpassing GRPO by 1.4\%/1.6\% and HintGRPO by 3.2\%/3.9\%, confirming its effectiveness at the 32B scale.
\subsection{Additional Supervised Baselines}
\label{Appendix:B.SFT}
Table~\ref{tab:sft_variants} compares ADHint with SFT$^{\dagger}$, a more comparable supervised baseline that conditions on the prompt and hint prefix to generate the continuation. SFT$^{\dagger}$ $\rightarrow$ RL further applies GRPO after this supervised stage.
SFT$^{\dagger}$ and SFT$^{\dagger}$ $\rightarrow$ RL are partly weaker than their standard counterparts, indicating that supervised learning from a single off-policy trajectory does not reliably improve generalization.
ADHint achieves the best overall performance at both 3B and 7B scales, demonstrating the advantage of difficulty-aware hint guidance in balancing on-policy exploration and off-policy imitation.
\subsection{Computational Overhead}
\label{Appendix:B.overhead}
ADHint and GRPO both generate 16 rollouts per query: ADHint generates eight naive-rollouts followed by eight hint-rollouts, whereas GRPO generates 16 naive-rollouts.
Since AH-SDP, AE-RDP, and CGM involve only basic matrix/statistical operations with $O(1)$ complexity, latency mainly comes from rollout generation.
As a result, ADHint averages 52.8 seconds per step, compared with 41.3 seconds for GRPO.
On the same hardware, training takes 8.6 hours for ADHint, 6.9 hours for GRPO, 15.5 hours for LUFFY, 26.4 hours for StepHint, 9.3 hours for HintGRPO, and 21.2 hours for GHPO.
Thus, ADHint incurs only limited additional overhead over GRPO while remaining more efficient than the other hint-based baselines.
\begin{table*}[htbp]
	\centering
	\captionsetup{skip=4pt}
	\renewcommand{\arraystretch}{1.0}
	\footnotesize
	\setlength{\tabcolsep}{4.0pt}
	\begin{tabular}{@{}lccccccc@{}}
		\toprule[1pt]
		\textbf{Method} & MathVista & MathVerse & We-Math & MMMU & MMMU-Pro & LogicVista & \textbf{Avg.} \\
		\midrule[0.8pt]
		
		GRPO~\cite{shao2024deepseekmath} & 76.8/77.6 & 75.9/77.3 & 78.6/79.7 & \underline{64.5}/\underline{65.9} & 50.6/51.2 & \textbf{63.5}/60.0 & 68.3/\underline{68.6} \\ 
		LUFFY~\cite{yan2025learning} & 70.2/68.1 & 64.5/68.7 & 60.7/67.5 & 48.2/61.4 & 35.2/47.4 & 41.7/53.7 & 53.4/61.1 \\ 
		StepHint~\cite{zhang2025stephint} & 72.3/67.9 & 64.5/67.8 & 63.2/66.6 & 36.2/43.3 & 37.0/47.2 & 43.3/54.6 & 52.8/57.9 \\ 
		HintGRPO~\cite{Huang_2025_ICCV} & \textbf{80.4}/\underline{79.0} & \textbf{79.0}/\underline{77.8} & \underline{81.4}/\underline{79.9} & 57.6/56.6 & \underline{53.0}/\underline{52.1} & 59.7/\underline{60.5} & \underline{68.5}/67.7 \\ 
		GHPO~\cite{liu2025ghpo} & 78.8/76.4 & 72.7/61.7 & 78.7/74.6 & 63.7/60.4 & 50.8/46.6 & \underline{61.8}/56.4 & 67.8/62.7 \\ 
		
		\midrule[0.8pt]
		\rowcolor{green!15} 
		\textbf{ADHint (Ours)} & \underline{79.5}/\textbf{79.6} & \underline{77.8}/\textbf{78.1} & \textbf{81.6}/\textbf{81.6} & \textbf{68.3}/\textbf{68.7} & \textbf{53.7}/\textbf{54.1} & \underline{61.8}/\textbf{62.4} & \textbf{70.5}/\textbf{70.8} \\
		
		\bottomrule[1pt]
	\end{tabular}
	\caption{
		\textbf{Comparative performance on MiMo-VL-7B-SFT-2508.}
		For each benchmark, we report \textit{pass@1} (the first value) and \textit{avg@8} (the second value).
	}
	\label{tab:3}
	\vspace{-1.em}
\end{table*}
\begin{table*}[htbp]
	\centering
	\captionsetup{skip=4pt}
	\renewcommand{\arraystretch}{1.0}
	\footnotesize
	\setlength{\tabcolsep}{14.0pt}
	\begin{tabular}{@{}lcccccc@{}}
		\toprule[1pt]
		\textbf{Method} & AIME & AMC & Minerva & MATH500 & Olympiad & \textbf{Avg.} \\
		\midrule[0.8pt]
		GRPO~\cite{shao2024deepseekmath}$^{\dagger}$ & 27.0 & 62.5 & 34.6 & 81.0 & 44.8 & 50.0 \\
		LUFFY~\cite{yan2025learning} & 31.1 & 67.5 & 37.9 & 81.8 & 44.0 & 52.5 \\
		StepHint~\cite{zhang2025stephint} & 29.6 & 61.4 & 36.4 & 81.0 & 40.4 & 49.8 \\
		HintGRPO~\cite{Huang_2025_ICCV} & \underline{32.8} & \underline{71.1} & 35.3 & \underline{83.0} & 44.7 & 53.4 \\
		GHPO~\cite{liu2025ghpo}$^{\dagger}$ & 32.0 & 70.0 & \underline{38.2} & 82.2 & \underline{45.3} & \underline{53.5} \\
		\midrule[0.8pt]
		\rowcolor{green!15}
		\textbf{ADHint (Ours)} & \textbf{33.6} & \textbf{72.3} & \textbf{41.5} & \textbf{84.6} & \textbf{47.3} & \textbf{55.9} \\
		\bottomrule[1pt]
	\end{tabular}
	\caption{\textbf{Math reasoning performance on Qwen2.5-Math-7B.} We use the same training data and experimental settings as GHPO~\cite{liu2025ghpo}. We report \textit{avg@32} on AIME and \textit{pass@1} on all other benchmarks. $^{\dagger}$ Results are taken from GHPO.}
	\label{tab:4}
	\vspace{-1.2em}
\end{table*}
\begin{table}[htbp]
	\centering
	\captionsetup{skip=4pt}
	\renewcommand{\arraystretch}{1.0}
	\footnotesize
	\setlength{\tabcolsep}{1.pt}
	\begin{tabular}{@{}>{\raggedright\arraybackslash}p{0.74\columnwidth}
			>{\centering\arraybackslash}p{0.20\columnwidth}@{}}
		\toprule[1pt]
		\textbf{Method} & PMC-Test \\
		\midrule[0.8pt]
		GRPO~\cite{shao2024deepseekmath} & 53.4 \\
		LUFFY~\cite{yan2025learning} & 49.4 \\
		StepHint~\cite{zhang2025stephint} & 48.3 \\
		HintGRPO~\cite{Huang_2025_ICCV} & 20.9 \\
		GHPO~\cite{liu2025ghpo} & 52.9 \\
		\midrule[0.8pt]
		\rowcolor{green!15}
		\textbf{ADHint (Ours)} & \textbf{55.1} \\
		\bottomrule[1pt]
	\end{tabular}
	\caption{\textbf{Medical VQA performance on PMC-Test~\cite{zhang2023pmc} using Qwen2.5-VL-7B.}}
	\label{tab:5}
	\vspace{-2.0em}
\end{table}
\section{Detailed Related Work}
\label{Appendix:C}
\begin{table*}[t]
	\centering
	\captionsetup{skip=4pt}
	\renewcommand{\arraystretch}{1.0}
	\footnotesize
	\setlength{\tabcolsep}{10.0pt}
	\begin{tabular}{@{}lccccccc@{}}
		\toprule[1pt]
		\textbf{Setting} 
		& MathVista & MathVerse & We-Math & MMMU & MMMU-Pro & LogicVista & \textbf{Avg.} \\
		\midrule[0.8pt]
		
		$w_{\text{max}}=0$ 
		& 73.4/73.4 & 55.3/59.2 & 69.8/68.8 & 55.3/55.6 & 40.4/40.0 & 44.9/46.4 & 56.5/57.2 \\
		
		\rowcolor{green!15}
		\textbf{$w_{\text{max}}=0.2$} 
		& \textbf{74.4}/\textbf{74.8} 
		& 60.6/\textbf{61.8} 
		& \textbf{69.9}/\textbf{70.8} 
		& \textbf{56.4}/\textbf{56.4} 
		& \textbf{41.3}/\textbf{41.8} 
		& \textbf{48.7}/\textbf{49.4} 
		& \textbf{58.6}/\textbf{59.2} \\
		
		$w_{\text{max}}=0.4$ 
		& 73.8/73.5 
		& \textbf{62.6}/57.9 
		& 68.2/67.5 
		& 54.8/54.5 
		& \textbf{41.3}/40.1 
		& 47.3/47.5 
		& 58.0/56.8 \\
		
		$w_{\text{max}}=0.6$ 
		& 73.8/72.4 
		& 56.9/59.7 
		& 65.5/67.7 
		& 54.9/55.5 
		& 39.3/40.2 
		& 46.9/47.5 
		& 56.2/57.2 \\
		
		\bottomrule[1pt]
	\end{tabular}
	\caption{
		\textbf{Effect of the maximum hint ratio $w_{\text{max}}$.}
		For each benchmark, we report \textit{pass@1} (the first value) and \textit{avg@8} (the second value).
	}
	\label{tab:7}
	\vspace{-1.2em}
\end{table*}
\begin{table*}[t]
	\centering
	\captionsetup{skip=4pt}
	\renewcommand{\arraystretch}{1.0}
	\footnotesize
	\setlength{\tabcolsep}{10.0pt}
	\begin{tabular}{@{}lccccccc@{}}
		\toprule[1pt]
		\textbf{Setting} 
		& MathVista & MathVerse & We-Math & MMMU & MMMU-Pro & LogicVista & \textbf{Avg.} \\
		\midrule[0.8pt]
		
		$\alpha=0.25$ 
		& 72.0/72.8 & 56.0/59.5 & 67.9/68.6 & 55.3/55.6 & 40.9/40.7 & \textbf{48.9}/48.1 & 56.8/57.6 \\
		
		\rowcolor{green!15}
		\textbf{$\alpha=0.5$} 
		& 74.4/74.8 & \textbf{60.6}/61.8 & \textbf{69.9}/\textbf{70.8} & \textbf{56.4}/\textbf{56.4} & \textbf{41.3}/\textbf{41.8} & 48.7/\textbf{49.4} & \textbf{58.6}/\textbf{59.2} \\
		
		$\alpha=0.75$ 
		& \textbf{76.0}/\textbf{75.1} & 53.4/\textbf{62.8} & 66.3/68.6 & 55.2/56.3 & 40.3/41.2 & 47.5/47.4 & 56.5/58.6 \\
		
		\bottomrule[1pt]
	\end{tabular}
	\caption{
		\textbf{Effect of the threshold $\alpha$ in CGM.} 
		 For each benchmark, we report \textit{pass@1} (the first value) and \textit{avg@8} (the second value).
	}
	\label{tab:8}
	\vspace{-0.6em}
\end{table*}
\subsection{Multimodal Reinforcement Learning}
GRPO~\cite{shao2024deepseekmath} removes the critic network from PPO~\cite{schulman2017proximal} and estimates advantages through intra-group sampling. This approach has become a cornerstone for the development of recent reasoning models~\cite{yang2025qwen3, zeng2025glm, team2025kimi}. 
In parallel, the multimodal community has explored various strategies to enhance the reasoning ability of MLLMs. Some works refine the reasoning style by encouraging models to ``think with images''~\cite{zheng2025deepeyes, su2025pixel}, explicitly grounding reasoning in visual features and directing attention to specific objects or regions. VL-ReThinker~\cite{wang2025vl} introduces an explicit reflection phase into the reasoning process, allowing the model to revisit and revise its initial thoughts. Other methods inject noise into rollouts~\cite{liu2025noisyrollout, wang2025perception} to improve training robustness and mitigate hallucinations. There are also approaches that design more precise reward functions for tasks such as object detection and semantic segmentation~\cite{liu2025visionreasoner, liu2025visual}, as well as methods that select or curate training data to improve efficiency~\cite{liang2025modomodomultidomaindatamixtures}. However, these advances primarily unlock the latent potential of pretrained models, often failing to expand their fundamental capability boundaries.
\subsection{Reinforcement Learning with Hints}
Recent work has investigated how to incorporate off-policy data into RL-based post-training. Some approaches directly integrate off-policy data into the SFT or RL objective~\cite{yan2025learning,ma2025learning,dong2025rl,fu2025srft}, but they struggle with the substantial distribution mismatch between off-policy data and the current policy model. To alleviate this issue, some methods design corrections for importance sampling~\cite{yan2025learning, dong2025rl}, control the trade-off between SFT and RL losses via entropy-based schedules~\cite{fu2025srft}, or alternate between RL and SFT phases~\cite{ma2025learning}. Hint-based methods address this problem more flexibly by guiding the model's reasoning with partial hints extracted from off-policy data while preserving its exploration behavior. UFT~\cite{liu2025uft} and Prefix-RFT~\cite{huang2025blending} employ the same temporally annealed hint ratios for all samples. SEELE~\cite{li2025staying} introduces learnable parameters to predict instance-specific hint ratios in a multi-round training scheme. Other methods reserve hints only for the hardest samples and determine the hint ratio from a fixed, predefined set via uniform sampling~\cite{chen2025data}, binary search~\cite{zhang2025bread}, or progressive search~\cite{Huang_2025_ICCV, liu2025ghpo}. StepHint~\cite{zhang2025stephint} samples rollouts under multiple hint ratios and merges the resulting hint-rollouts, naive-rollouts, and ground truth off-policy data into a single group for advantage estimation. However, these methods often overlook difficulty as a key factor, leading to insufficient and unstable training. In this work, we propose ADHint, which explicitly incorporates the sample difficulty prior into the hint-ratio schedule and the rollout difficulty posterior into relative-advantage estimation. This difficulty-aware design stabilizes training, enhances exploration, and expands the model's knowledge boundary when learning from off-policy hints. 
\begin{table*}[htbp]
	\centering
	\captionsetup{skip=4pt}
	\renewcommand{\arraystretch}{1.0}
	\footnotesize
	\setlength{\tabcolsep}{4.0pt}
	\begin{tabular}{@{}lccccccc@{}}
		\toprule[1pt]
		\textbf{Method} & MathVista & MathVerse & We-Math & MMMU & MMMU-Pro & LogicVista & \textbf{Avg.} \\
		\midrule[0.8pt]
		
		GRPO~\cite{shao2024deepseekmath}
		& 77.4/78.0
		& \textbf{67.7}/67.7
		& 78.9/78.7
		& 66.3/65.8
		& 51.7/51.0
		& 58.3/\textbf{58.3}
		& 66.7/66.6 \\
		
		HintGRPO~\cite{Huang_2025_ICCV}
		& 75.7/76.4
		& 67.0/64.8
		& 75.1/75.2
		& 64.8/64.5
		& 49.1/48.8
		& 57.8/56.2
		& 64.9/64.3 \\
		
		\midrule[0.8pt]
		\rowcolor{green!15}
		\textbf{ADHint (Ours)}
		& \textbf{79.6}/\textbf{80.9}
		& 66.7/\textbf{68.5}
		& \textbf{80.4}/\textbf{80.5}
		& \textbf{68.3}/\textbf{68.2}
		& \textbf{53.9}/\textbf{53.0}
		& \textbf{59.5}/58.0
		& \textbf{68.1}/\textbf{68.2} \\
		
		\bottomrule[1pt]
	\end{tabular}
	\caption{
		\textbf{Comparative performance on Qwen2.5-VL-32B.}
		For each benchmark, we report \textit{pass@1} (the first value) and \textit{avg@8} (the second value).
	}
	\label{tab:32b}
	\vspace{-1.4em}
\end{table*}
\section{Detailed Experimental Settings}
\label{Appendix:D}
\subsection{Training data}
\label{Appendix:D.1}
\begin{figure*}[t]
	\centering

	\begin{subfigure}[t]{0.46\linewidth}
		\centering
		\includegraphics[width=\linewidth]{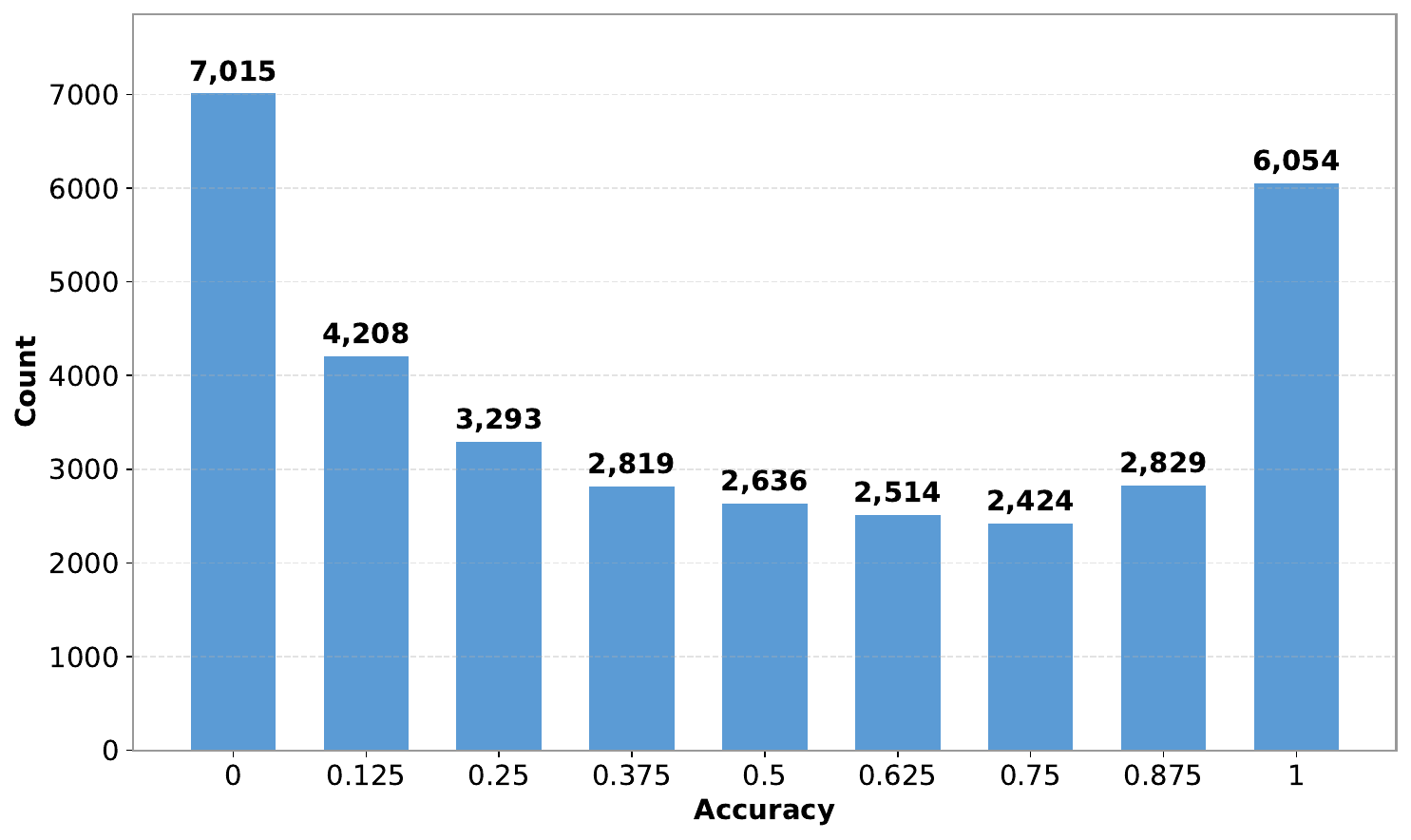}
		\caption{\textit{Avg@8} distribution of Qwen2.5-VL-7B (base model) on the 34K filtered samples.}
		\label{distribution:a}
	\end{subfigure}
	\hfill
	\begin{subfigure}[t]{0.46\linewidth}
		\centering
		\includegraphics[width=\linewidth]{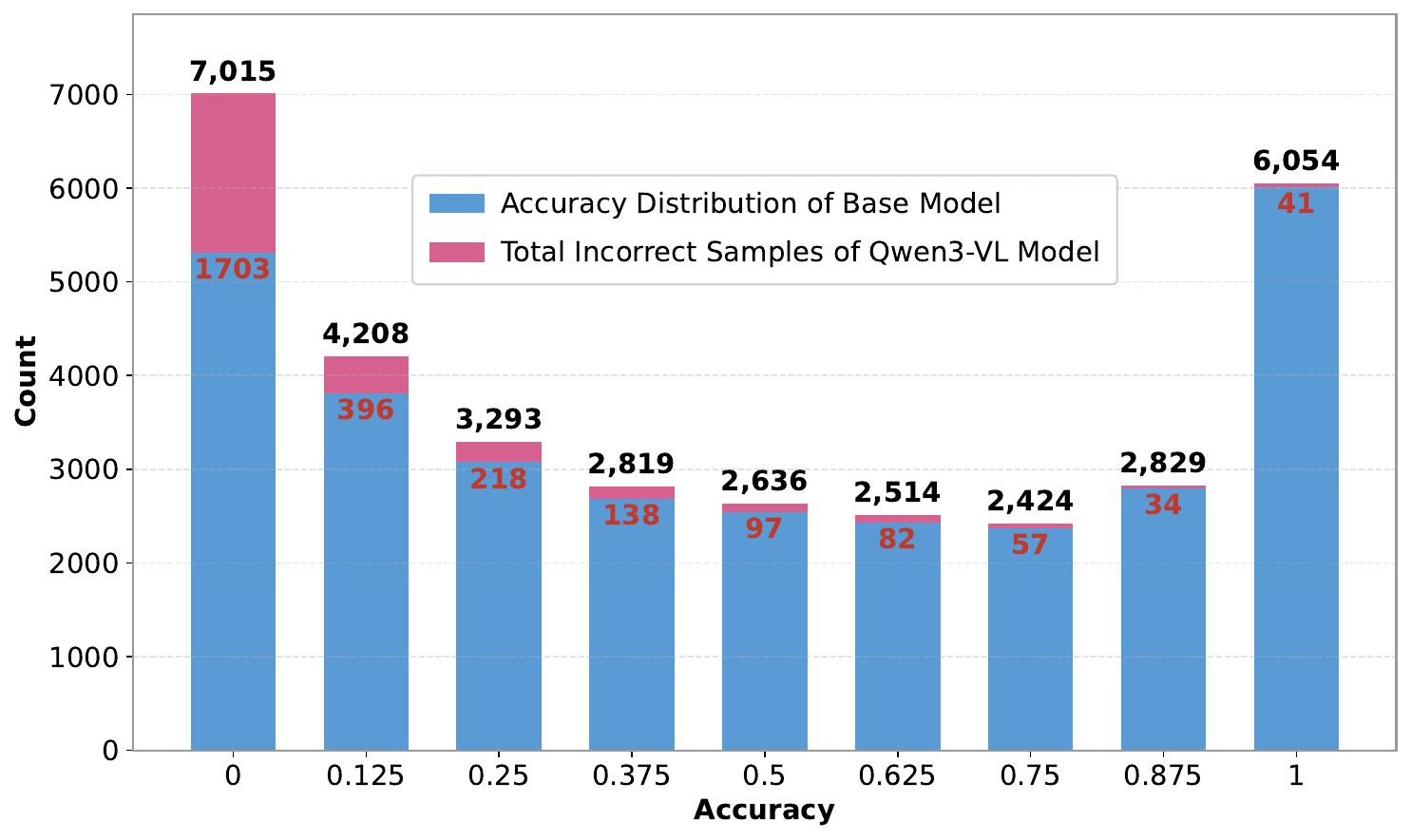}
		\caption{Distribution of samples for which Qwen3-VL-30B-A3B-Thinking fails to produce any correct hint among eight candidates.}
		\label{distribution:b}
	\end{subfigure}
	
	\vspace{4mm}

	\begin{subfigure}[t]{0.46\linewidth}
		\centering
		\includegraphics[width=\linewidth]{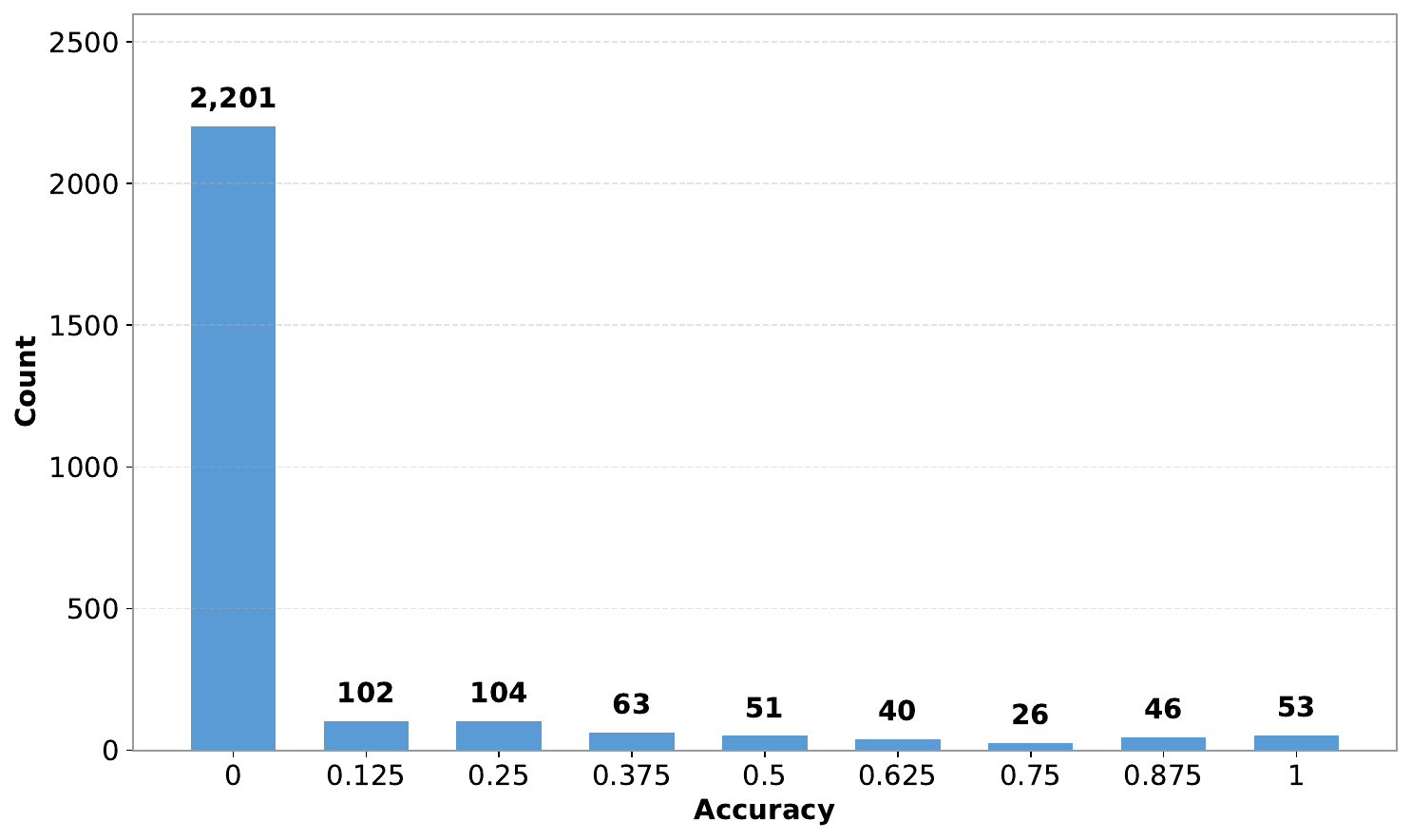}
		\caption{\textit{Avg@8} distribution of Qwen3-VL-235B-A22B-Thinking on the remaining hard samples.}
		\label{distribution:c}
	\end{subfigure}
	\hfill
	\begin{subfigure}[t]{0.46\linewidth}
		\centering
		\includegraphics[width=\linewidth]{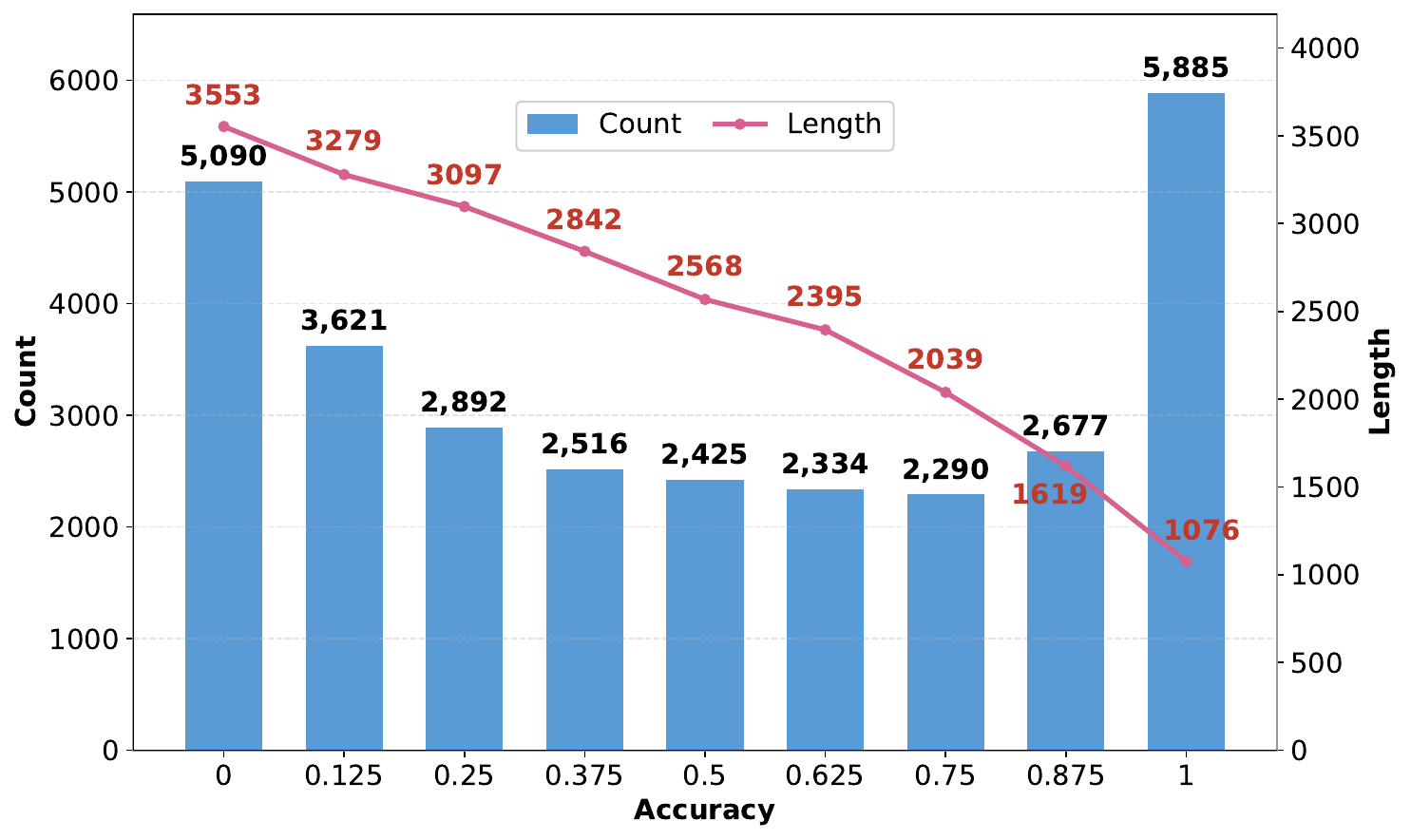}
		\caption{Hint-length distribution and base model accuracy after hint sampling.}
		\label{distribution:d}
	\end{subfigure}
	
	\caption{
		\textbf{Accuracy and length statistics during ViRL-Hint17k construction.} 
		At each sampling stage, we independently generate eight reasoning trajectories with temperature 1.0.
	}
	\label{distribution}
	\vspace{-0.6em}
\end{figure*}

For MLLM training, we construct an off-policy hint dataset based on ViRL39K~\cite{wang2025vl}, which covers STEM, chart reasoning, document reasoning, and spatial reasoning. We first remove samples that involve multi-image reasoning, text-only reasoning, or format errors, obtaining roughly 34K examples. Figure~\ref{distribution:a} shows the \textit{avg@8} distribution of Qwen2.5-VL-7B on this filtered set. We then sample hints with Qwen3-VL-30B-A3B-Thinking, and Figure~\ref{distribution:b} reports the distribution of examples for which this model fails to produce any correct hint among eight candidates. For these hard cases, we further sample hints with Qwen3-VL-235B-A22B-Thinking, whose \textit{avg@8} distribution is shown in Figure~\ref{distribution:c}. Among samples with at least one correct hint, we discard those with hint length exceeding 8K tokens and randomly select one of the eight candidates as the final hint, with the resulting hint-length distribution shown in Figure~\ref{distribution:d}. Finally, we remove samples that are excessively simple for Qwen2.5-VL-7B and obtain ViRL-Hint17k, which we will open-source. 

For Medical VQA, we similarly employ Qwen3-VL-30B-A3B-Thinking to sample hints from the PMC-VQA training set~\cite{zhang2023pmc}, ultimately constructing a filtered subset of 15K samples. For LLM training, we utilize NuminaMath-S curated by GHPO~\cite{liu2025ghpo}, which comprises 18.3K samples combined from MATH~\cite{hendrycks2measuring} and NuminaMath-1.5~\cite{li2024numinamath}.
\begin{table*}[htbp]
	\centering
	\captionsetup{skip=4pt}
	\renewcommand{\arraystretch}{1.0}
	\footnotesize
	\setlength{\tabcolsep}{8.5pt}
	\begin{tabular}{@{}lccccccc@{}}
		\toprule[1pt]
		\textbf{Method} & MathVista & MathVerse & We-Math & MMMU & MMMU-Pro & LogicVista & \textbf{Avg.} \\
		\midrule[0.8pt]
		
		\rowcolor{gray!10}
		\multicolumn{8}{c}{\textit{Base MLLM: Qwen2.5-VL-3B}} \\
		
		SFT
		& 51.9/61.3
		& 31.2/\underline{54.2}
		& 32.5/58.5
		& 28.6/46.3
		& 17.1/29.4
		& 18.1/37.8
		& 29.9/47.9 \\
		
		SFT $\rightarrow$ RL
		& 53.6/\underline{63.3}
		& 37.4/54.1
		& 38.3/\textbf{60.6}
		& 30.6/47.7
		& 18.7/\underline{30.9}
		& 18.1/39.9
		& 32.8/\underline{49.4} \\
		
		SFT$^{\dagger}$
		& 30.6/43.7
		& 14.6/29.6
		& 21.4/28.6
		& 30.2/36.4
		& 22.5/27.0
		& 21.6/29.3
		& 23.5/32.4 \\
		
		SFT$^{\dagger}$ $\rightarrow$ RL
		& 55.5/62.5
		& 40.8/\textbf{55.8}
		& 47.0/57.2
		& 38.7/48.3
		& 24.3/30.5
		& 27.9/\underline{40.0}
		& 39.0/49.1 \\
		
		GRPO
		& \underline{63.9}/63.1
		& \underline{43.1}/45.0
		& \underline{59.0}/59.0
		& \underline{47.5}/\underline{49.1}
		& \underline{30.7}/30.1
		& \textbf{43.8}/38.9
		& \underline{48.0}/47.5 \\
		
		\midrule[0.8pt]
		\rowcolor{green!15}
		\textbf{ADHint (Ours)}
		& \textbf{64.8}/\textbf{64.3}
		& \textbf{49.5}/50.6
		& \textbf{60.4}/\underline{59.1}
		& \textbf{51.4}/\textbf{50.0}
		& \textbf{33.3}/\textbf{32.0}
		& \underline{42.6}/\textbf{41.4}
		& \textbf{50.3}/\textbf{49.6} \\
		
		\midrule[0.8pt]
		\rowcolor{gray!10}
		\multicolumn{8}{c}{\textit{Base MLLM: Qwen2.5-VL-7B}} \\
		
		SFT
		& 59.0/68.7
		& 40.1/64.0
		& 44.1/70.1
		& 31.2/51.0
		& 20.6/35.6
		& 23.7/46.0
		& 36.5/55.9 \\
		
		SFT $\rightarrow$ RL
		& 64.7/71.1
		& \underline{56.4}/\textbf{67.2}
		& 54.1/\textbf{72.5}
		& 35.9/54.2
		& 25.8/38.3
		& 28.8/\underline{47.8}
		& 44.3/\underline{58.5} \\
		
		SFT$^{\dagger}$
		& 52.2/57.6
		& 27.1/41.4
		& 34.0/44.9
		& 36.8/44.2
		& 24.2/29.5
		& 33.9/39.2
		& 34.7/42.8 \\
		
		SFT$^{\dagger}$ $\rightarrow$ RL
		& 62.6/69.8
		& 49.3/\underline{64.1}
		& 52.5/70.5
		& 41.7/52.8
		& 27.9/37.2
		& 33.0/46.3
		& 44.5/56.8 \\
		
		GRPO
		& \underline{73.4}/\underline{73.4}
		& 55.3/59.2
		& \underline{69.8}/68.8
		& \underline{55.3}/\underline{55.6}
		& \underline{40.4}/\underline{40.0}
		& \underline{44.9}/46.4
		& \underline{56.5}/57.2 \\
		
		\midrule[0.8pt]
		\rowcolor{green!15}
		\textbf{ADHint (Ours)}
		& \textbf{74.4}/\textbf{74.8}
		& \textbf{60.6}/61.8
		& \textbf{69.9}/\underline{70.8}
		& \textbf{56.4}/\textbf{56.4}
		& \textbf{41.3}/\textbf{41.8}
		& \textbf{48.7}/\textbf{49.4}
		& \textbf{58.6}/\textbf{59.2} \\
		
		\bottomrule[1pt]
	\end{tabular}
	\caption{
		\textbf{Comparison with additional supervised baselines on Qwen2.5-VL models.}
		SFT$^{\dagger}$ is trained to generate the continuation given the prompt and hint prefix. For each benchmark, we report \textit{pass@1} (the first value) and \textit{avg@8} (the second value).
	}
	\label{tab:sft_variants}
\end{table*}
\subsection{Benchmarks}
\label{Appendix:D.2}
For MLLM evaluation, we employ six benchmarks that cover mathematical reasoning, multidisciplinary multimodal reasoning, and logical reasoning:
\begin{itemize}
    \item \textbf{MathVista} --- A consolidated benchmark for mathematical reasoning in visual contexts, which unifies math-focused and vision–language tasks into a single suite and requires fine-grained visual understanding and compositional reasoning. We use the testmini split for evaluation.
    \item \textbf{MathVerse} --- A holistic visual math benchmark that is crafted to test whether multimodal models truly understand mathematical diagrams, providing multiple variants of each problem with different modality configurations to probe robust diagram understanding. We use the Vision-Only subset.
    \item \textbf{We-Math} --- A large-scale benchmark that investigates human-like mathematical reasoning in MLLMs, decomposing problems into sub-problems and introducing multi-dimensional metrics to analyze knowledge acquisition, generalization, and rote memorization behaviours. We use the testmini split.
    \item \textbf{MMMU} --- A massive multi-discipline multimodal understanding benchmark that spans college-level subjects such as science, engineering, humanities, and business, and demands deliberate integration of visual and textual information for complex problem solving. We use the validation split.
    \item \textbf{MMMU-Pro} --- A more challenging variant of MMMU that filters out text-only questions, augments candidate options, and introduces vision-only settings, thereby providing a more robust assessment of genuine multimodal understanding and reasoning. We use the standard subset with 10 options and the vision subset to compute the overall score.
    \item \textbf{LogicVista} --- A benchmark that targets integrated logical reasoning in visual environments, focusing on compositional, step-by-step reasoning over diagrams and structured visual scenes.
\end{itemize}

For Medical VQA, we evaluate on the PMC-VQA test set, a comprehensive benchmark derived from PubMed Central that assesses a model's ability to comprehend medical images and generate clinically grounded answers.

For LLM evaluation, we employ five mathematical reasoning benchmarks that align with our training domain:
\begin{itemize}
    \item \textbf{AIME24} --- A dataset that contains 30 competition-style problems from the 2024 American Invitational Mathematics Examination (AIME), a well-known high-school contest featuring challenging short-answer problems.
    \item \textbf{AMC} --- A benchmark constructed from the American Mathematics Competitions, covering a broad range of high-school mathematics topics and providing multiple-choice questions that emphasize conceptual understanding and problem-solving skills.
    \item \textbf{MATH500} --- A high-difficulty math competition benchmark containing 500 problems across multiple subjects, designed to assess advanced problem-solving and multi-step symbolic reasoning.
    \item \textbf{Minerva} --- A quantitative reasoning benchmark with around 500 carefully curated math problems that require multi-step derivations and the integration of textual descriptions with formal mathematical manipulations.
    \item \textbf{OlympiadBench} --- An Olympiad-level benchmark that comprises thousands of mathematically and scientifically challenging problems, offering a rigorous testbed for models' higher-order reasoning and abstraction abilities.
\end{itemize}
\subsection{Implementation Details}
\label{Appendix:D.3}
As a supplement to the main text, Tables~\ref{tab:grpo config} and~\ref{tab:SFT config} summarize the training configs of the GRPO and SFT baselines, respectively. Table~\ref{tab:ADHint config} reports the default configs for ADHint, with the exception of MiMo-VL-7B-SFT-2508, Qwen2.5-Math-7B, and Medical VQA (based on Qwen2.5-VL-7B), where the maximum hint ratio $w_{\text{max}}$ is set to 0.4. Note that both GRPO and ADHint use 16 rollouts to ensure a fair comparison.
\subsection{Baselines}
\label{Appendix:D.4}
We compare ADHint against a diverse set of baselines to validate its effectiveness. 
These include standard on-policy RL, supervised fine-tuning, and recent methods that attempt to incorporate off-policy data into on-policy RL training:
\begin{itemize}
    \item \textbf{GRPO} --- an on-policy RL method that estimates group-level relative advantages from rollouts and relies solely on an outcome reward function or model.
    \item \textbf{SFT} --- a supervised baseline that directly trains the model to imitate complete off-policy trajectories (full hints in our setting), which encourages memorization of both knowledge and response style.
    \item \textbf{SFT $\rightarrow$ RL} --- a common cold-start strategy that first performs SFT and then applies RL, serving as a strong baseline that combines the benefits of SFT and RL.
    \item \textbf{LUFFY} --- a method that augments GRPO by incorporating off-policy reasoning trajectories into the GRPO objective.
    \item \textbf{StepHint} --- a hint-based method that mixes hint-rollouts under multiple hint ratios, naive-rollouts, and off-policy reasoning trajectories into a single GRPO group for advantage estimation.
    \item \textbf{HintGRPO} --- a method that focuses on the hardest samples where the current policy fails to obtain a positive rollout. It progressively increases the hint ratio ($1/3$, $2/3$) until a positive rollout is found, followed by a one-step policy update.
    \item \textbf{GHPO} --- a method that treats hints as part of the query and excludes hint tokens from loss computation. It similarly increases the hint ratio (25\%, 50\%, 75\%) until a positive rollout is found.
\end{itemize}

For methods that do not support MLLM training or whose code is not publicly available, we carefully reproduce them within the EasyR1 framework~\cite{zheng2025easyr1}. 
We note that several baselines exhibit instability and crash during training. 
We provide a detailed analysis in Section~\ref{Appendix:E}.
\begin{table}[t]
\centering
\footnotesize
\renewcommand{\arraystretch}{1.0}
\setlength{\tabcolsep}{2.5pt}
\begin{tabular}{@{}>{\bfseries}c l c@{}}
\toprule
\textbf{Model} & \textbf{Config} & \textbf{Value} \\
\midrule
\multirow{10}{*}{\parbox{2.0cm}{\centering Qwen2.5-VL\\\&\\Qwen3-VL\\\&\\MiMo-VL-SFT}} 
 & max prompt length & 4096 \\
 & max response length & 8192 \\
 & temperature & 1.0 \\
 & learning rate & 1e-6 \\
 & rollout batchsize & 128 \\
 & global batchsize & 64 \\
 & num rollouts & 16 \\
 & epoch & 2 \\
 & min pixels & $256 * 28 * 28$ \\
 & max pixels & $1280 * 28 * 28$ \\
\midrule
\multirow{9}{*}{Qwen2.5-Math} 
 & max prompt length & 2048 \\
 & max response length & 2048 \\
 & temperature & 1.0 \\
 & learning rate & 1e-6 \\
 & lr schedule & cosine \\
 & warmup ratio & 0.1 \\
 & rollout batchsize & 128 \\
 & global batchsize & 32 \\
 & epoch & 5 \\
\bottomrule
\end{tabular}
\caption{\textbf{Training configs for GRPO baselines.}}
\label{tab:grpo config}
\end{table}
\begin{table}[ht]
    \centering
    \footnotesize
    \renewcommand{\arraystretch}{1.0}
    \setlength{\tabcolsep}{3pt}
    \begin{tabular}{c l c}
        \toprule[1pt]
        \textbf{Method} & \textbf{Config} & \textbf{Value} \\
        \midrule[0.8pt]
        \multirow{7}{*}{\textbf{SFT}} 
        & learning rate & 5e-5 \\
        & lr schedule & cosine \\
        & warmup ratio & 0.1 \\
        & batchsize & 64 \\
        & epoch & 2 \\
        & min pixels & $256 \times 28 \times 28$ \\
        & max pixels & $1280 \times 28 \times 28$ \\
        \bottomrule[1pt]
    \end{tabular}
    \caption{\textbf{Training configs for SFT baselines.}}
    \label{tab:SFT config}
\end{table}
\begin{table}[ht]
    \centering
    \footnotesize
    \renewcommand{\arraystretch}{1.0}
    \setlength{\tabcolsep}{2pt}
    \begin{tabular}{@{}l l c@{}}
        \toprule[1pt]
        \textbf{Method} & \textbf{Config} & \textbf{Value} \\
        \midrule[0.8pt]
        \multirow{6}{*}{\textbf{ADHint}}
        & warmup steps & 5 \\
        & num of naive-rollouts $n$ & 8 \\
        & num of hint-rollouts $m$ & 8 \\
        & maximum hint ratio $w_{\text{max}}$ & 0.2 \\
        & sampling radius $R$ of the noise term $\sigma$ & 0.01 \\
        & threshold of CGM $\alpha$ & 0.5 \\
        \bottomrule[1pt]
    \end{tabular}
    \caption{\textbf{Training configs for ADHint.}}
    \label{tab:ADHint config}
\end{table}
\subsection{Reimplementation Details of Baselines}
\label{Appendix:D.5}
For all baselines, we strictly follow the official implementations and the settings reported in the original papers. 
We train each method for 2 epochs on the same ViRL-Hint17k dataset to ensure a fair comparison. 
Below, we describe the reimplementation details for each baseline. 
\paragraph{LUFFY.}
Following the paper, we add an off-policy objective to the standard GRPO objective to obtain the mixed-policy training. We also reimplement the policy shaping module to modulate the update gradients of off-policy tokens and set the shaping hyperparameter to the default value of 0.1 reported in LUFFY. We omit both the KL divergence constraint and the standard advantage normalization.
\paragraph{StepHint.}
We obtain hint-rollouts under hint ratios of 25\%, 50\%, and 75\%, and group them with naive-rollouts and complete off-policy reasoning trajectories into a single GRPO group for relative advantage estimation. 
We omit the KL constraint and mask hint tokens during optimization for negative hint-rollouts.
\paragraph{HintGRPO.}
We set the group number to $M=3$, so the hint ratio is progressively increased from 0 to $1/3$ and then $2/3$ until a positive rollout appears. 
We set the KL coefficient to 0.04 and the maximum response length to 8K tokens. This setting matches our ViRL-Hint17k dataset, which has a relatively long average hint length (around 3100 tokens) and is substantially longer than that in the original HintGRPO experiments. 
We do not reproduce the \textit{Text-Bias Calibration} module from HintGRPO because it is a test-time method for correcting perception bias and mitigating hallucinations rather than a training-time hint-based component.
\paragraph{GHPO.}
For the hardest samples where the current policy fails to produce a correct answer, we insert hints into the user prompt using the official prompt template. 
The hint ratio is progressively increased from 25\% to 50\% and then to 75\% until a positive rollout appears. We then use the corresponding hint-rollouts as the update signal for that sample. 
Following the original paper, we omit the KL constraint and apply a 5-step GRPO warmup. 
\section{Comparative Analysis of ADHint and Baselines}
\label{Appendix:E}
As shown in Figure~\ref{fig:2}, ADHint consistently outperforms all baselines. Here, we provide a detailed comparative analysis of ADHint and these baselines to better understand the sources of their performance differences.
\paragraph{SFT learns, RL generalizes.}
As shown in Table~\ref{tab:1}, compared with pure GRPO, the \mbox{$\mathrm{SFT} \rightarrow \mathrm{RL}$} pipeline improves \textit{avg@8} by 1.3\% but reduces \textit{pass@1} by 12.2\% at the 7B scale. This pattern indicates that SFT injects new knowledge into the model and raises the upper bound of its performance when multiple samples are allowed, whereas the subsequent RL phase, starting from an SFT-initialized policy, tends to lose part of the exploration capability that is crucial for strong \textit{pass@1} performance compared with directly training with on-policy RL. In practice, SFT encourages the model to follow a specific response style, which makes the response length and entropy during RL training concentrate in a relatively narrow range, thereby limiting exploration across diverse reasoning trajectories. In contrast, ADHint achieves the best results on both \textit{pass@1} and \textit{avg@8}. By explicitly accounting for difficulty, ADHint assigns an appropriate hint ratio to each sample and estimates fair relative advantages for each rollout, balancing exploration and imitation. As a result, the policy model simultaneously strengthens its reasoning ability and absorbs external knowledge from off-policy hints.
\paragraph{Training collapses when directly using full off-policy data.}
During our reimplementation of LUFFY~\cite{yan2025learning}, we observe that training collapses at an early stage (around 40 steps), accompanied by a sharp increase in both entropy and response length. We attribute this to over-imitation of the off-policy distribution, especially when there is a large discrepancy between the model that generates the off-policy data and the current policy. LUFFY feeds the complete off-policy trajectories into the mixed-policy GRPO objective as the off-policy term, while the other term remains the original on-policy GRPO objective. This forces the policy model to simultaneously fit two highly mismatched distributions, making training unstable and prone to collapse. In contrast, ADHint treats difficulty as a vital signal. It adjusts the hint ratio based on sample difficulty and estimates relative advantages based on rollout difficulty, achieving a better trade-off between learning from off-policy hints and maintaining exploratory on-policy reasoning. 
\paragraph{Fixed hint ratios always lead to poor generalization.}
GHPO~\cite{liu2025ghpo} and HintGRPO~\cite{Huang_2025_ICCV} share a similar design philosophy: they add hints with fixed ratios (e.g., 25\%, 50\%) only to the hardest problems. The key difference is that GHPO injects hints into the prompt and excludes them from the training loss, whereas HintGRPO treats hints as part of the response and includes them in the loss computation. In our experiments, we observe that GHPO exhibits large update magnitudes and highly unstable training, likely because the model is not truly learning the knowledge contained in the hints but instead adapting to an out-of-distribution reasoning pattern. In contrast, HintGRPO trains more stably and maintains a high average reward during training. However, as shown in Table~\ref{tab:1}, HintGRPO performs poorly on the benchmarks and shows limited generalization. The progressive hinting strategy for the hardest samples helps obtain positive rollouts, but it ignores sample difficulty and can provide excessive guidance, causing SFT-like overfitting to the hint distribution and weakening exploration. As a result, the model attains inflated training rewards under heavy hinting but fails to translate these gains into improvements on the benchmarks. By comparison, ADHint leverages a sample difficulty prior to assigning appropriate guidance to each example and uses low-variance update signals to gradually push the model's capability boundary, enabling it to learn from off-policy data while preserving its inherent exploration.
\paragraph{Imbalanced intra-group advantage estimation leads to abnormal updates.}
Methods like StepHint~\cite{zhang2025stephint} mix hint-rollouts with multiple fixed hint ratios, naive-rollouts, and ground-truth trajectories into a single group for advantage estimation without distinguishing among them. As shown in Figure~\ref{fig1:sub_b} of the main text, this imbalanced intra-group estimation allows longer hint-rollouts with more positive trajectories to receive disproportionately large advantages, which biases policy updates toward imitating hints and leads to abnormal update behavior. This issue becomes more severe on ViRL-Hint17k, where the average hint length is about 3{,}100 tokens, while Qwen2.5-VL models typically generate only 400–500 tokens. In contrast, ADHint introduces AE-RDP, which estimates rollout advantages using the difficulty posterior evaluated from naive-rollouts and hint-rollouts, balancing learning between the two rollout types. CGM and Selective Masking further modulate token-level gradients within hints, which promotes more stable and reliable policy updates.
\section{Case Study}
\label{Appendix:F}
\begin{figure*}[p]
    \centering
    \includegraphics[width=0.95\textwidth,height=0.78\textheight,keepaspectratio]{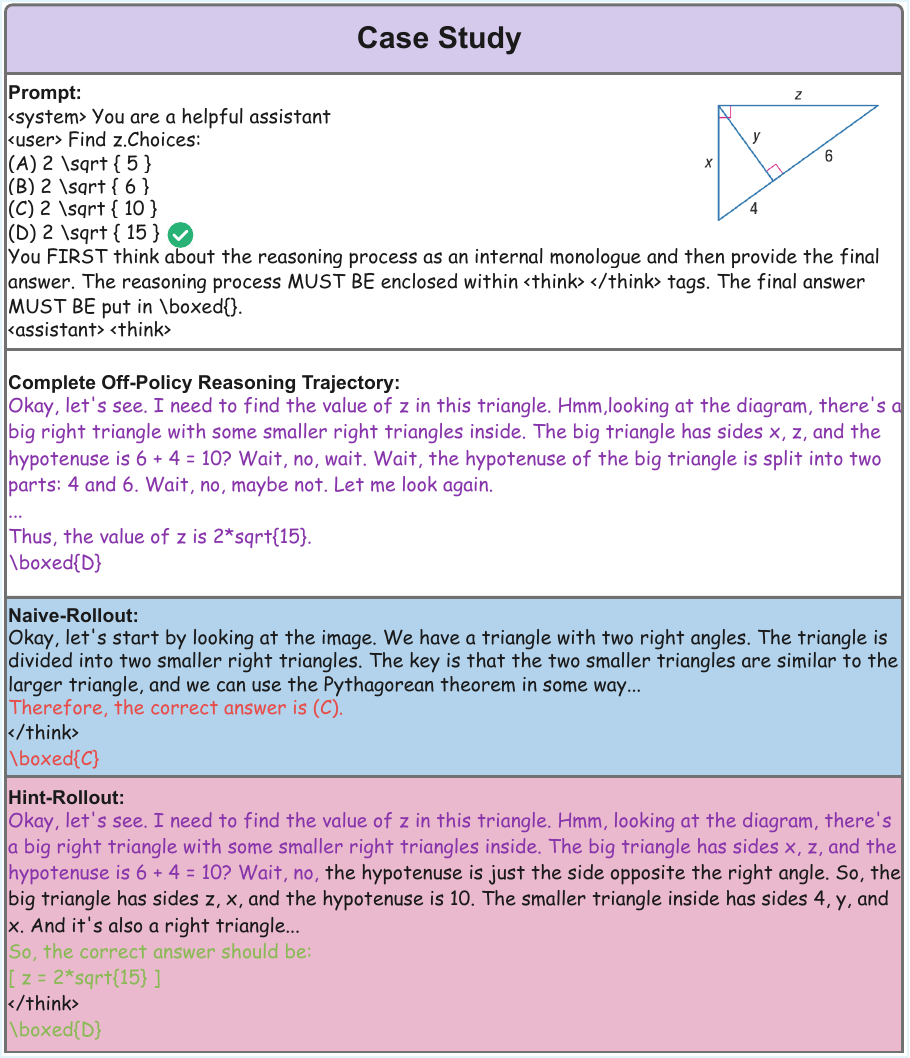}
    \caption{\textbf{A training example comparing naive-rollouts and hint-rollouts in ADHint.}}
    \label{fig:case1}
\end{figure*}

Figure~\ref{fig:case1} shows a representative training example. This case includes the extra prompt we added beyond the user query, which instructs the model to think before producing its response. For clarity, we omit this additional prompt in the subsequent cases. We see that the naive-rollout of the policy model fails to produce the correct answer. However, when we give the policy a hint taken from the complete off-policy reasoning trajectory, it continues the reasoning process and computes the final answer $z = 2\sqrt{15}$. In this way, the model not only learns the essential knowledge required to solve the problem from the hint but also improves its reasoning ability. 

\begin{figure*}[p]
    \centering
    \includegraphics[width=0.82\textwidth,height=0.82\textheight,keepaspectratio]{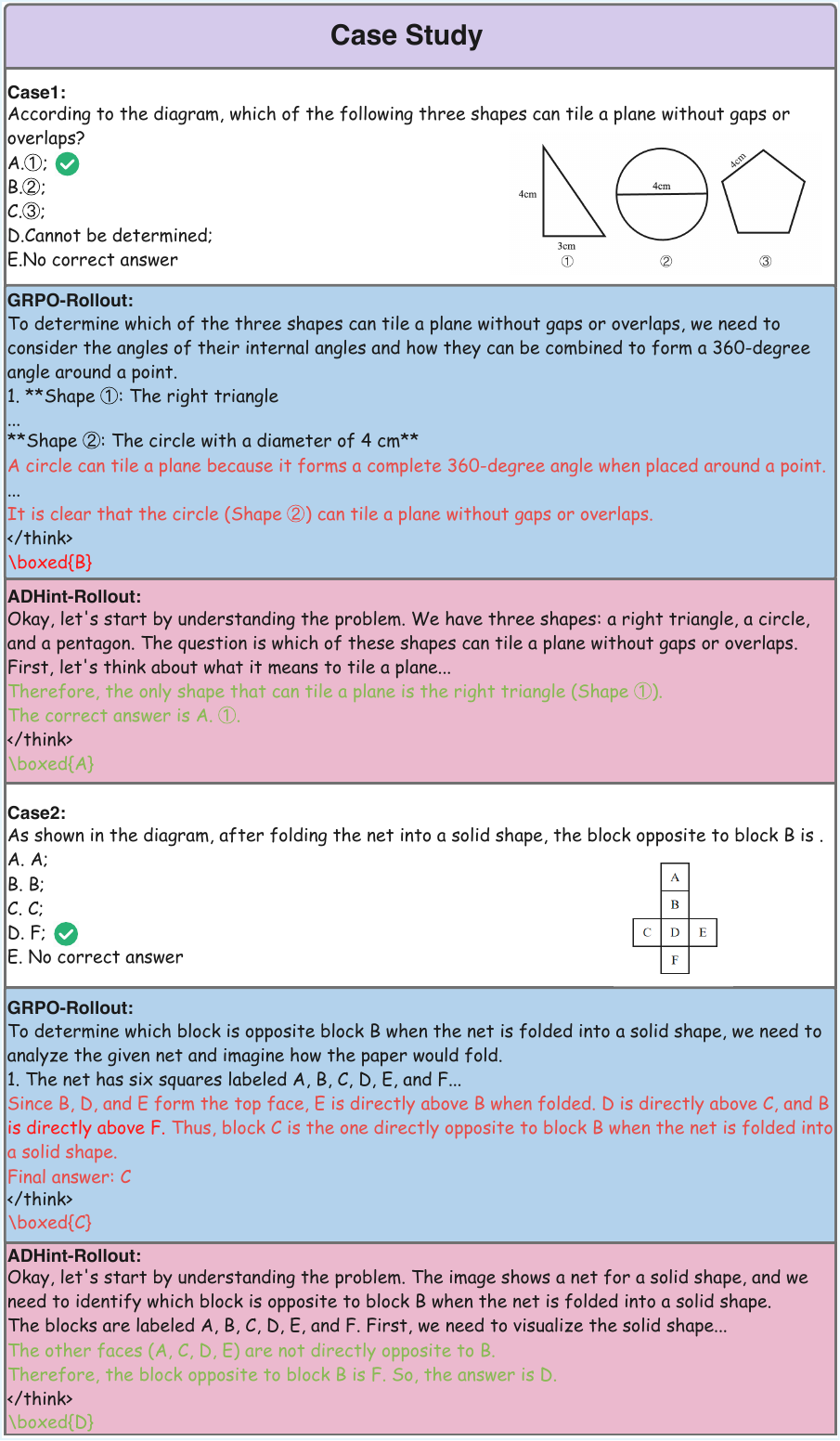}
    \caption{\textbf{Two benchmark examples comparing ADHint with the GRPO baseline.}}
    \label{fig:case2}
\end{figure*}

Figure~\ref{fig:case2} shows two representative benchmark examples. In the top case, the GRPO model lacks sufficient geometric knowledge and reaches an incorrect conclusion that \textit{``a circle can tile a plane because it forms a complete 360-degree angle when placed around a point.''} In contrast, the ADHint model performs careful reasoning: it first interprets the meaning of \textit{``tile a plane''} in the question, and then, through step-by-step reasoning, reaches the correct conclusion that \textit{``the only shape that can tile a plane is the right triangle.''} In the bottom case, the GRPO model produces flawed reasoning, such as claiming that \textit{``E is directly above B,''} and ultimately predicts the wrong option \textit{``C.''} By comparison, the ADHint model conducts logically coherent reasoning: it first \textit{``visualizes the solid shape''} and then gradually rules out incorrect options, finally selecting the correct answer \textit{``D.''} These case studies indicate that GRPO fails to acquire the necessary external knowledge and exhibits weak reasoning ability, whereas ADHint treats difficulty as a key factor, learns essential knowledge from hints in a controllable manner, and develops more structured logical reasoning. This demonstrates ADHint's strong ability to effectively integrate off-policy data into on-policy RL training.

\end{document}